\documentclass[mathematics,article,accept,pdftex,moreauthors]{Definitions/mdpi} 
\firstpage{1} 
\pubvolume{1}
\issuenum{1}
\articlenumber{0}
\pubyear{2025}
\copyrightyear{2025}
\externaleditor{Junlin Hu} 
\datereceived{16 September 2025 } 
\daterevised{17 December 2025 } 
\dateaccepted{27 December 2025 } 
\datepublished{ } 
\hreflink{https://doi.org/} 

	\usepackage[normalem]{ulem}
	\definecolor{flame}{rgb}{0.89, 0.35, 0.13}
    \definecolor{darkgreen}{rgb}{0,0.5,0}
	
	\usepackage{framed}

    \Title{Literary Language Mashup: Curating Fictions with Large Language Models}

	\Author{Gerardo Aleman Manzanarez 
 $^{1}$, Raul Monroy $^{1,}$*, Jorge Garcia Flores $^{2}$ and Hiram Calvo $^{3}$ 
	}
	
	\AuthorNames{Gerardo Aleman Manzanarez, Raul Monroy, Jorge Garcia Flores and Hiram Calvo}
	
	\address{%
		$^{1}$ \quad Tecnologico de Monterrey, Escuela de Ingenieria y Ciencias, Carretera al Lago de Guadalupe Km 3.5, Colonia Margarita Maza de Juarez 
, Atizapan de Zaragoza, C.P. 52926, 
, Mexico; 
		a00156749@tec.mx\\
        $^{2}$ \quad Laboratoire d’Informatique de Paris Nord, Centre National de la Recherche Scientifique, Université Sorbonne Paris Nord 
, 99 Av. Jean-Baptiste Clément, 93430 Villetaneuse, France; jgflores@lipn.univ-paris13.fr\\
        $^{3}$ \quad Centro de Investigación en Computación, Instituto Politécnico Nacional, Av. Juan de Dios Bátiz s/n,
		Col. Nueva Industrial Vallejo, 	Gustavo A. Madero C.P. 07738, 
, Mexico; hcalvo@cic.ipn.mx 
		}

	\corres{Correspondence: raulm@tec.mx}
	
	\abstract{
    {The artificial generation of text by computers has been a field of study in computer science since the beginning of the twentieth century, from Markov chains to Turing tests. This has evolved into automatic summarization and marketing chatbots.
The generation of literary texts by Large Language Models (LLMs) has also been an area of scholarly inquiry for over six decades. The literary quality of AI-generated text can be evaluated with GrAImes, an evaluation protocol grounded in literary theory and inspired by the editorial process of book publishers. This evaluation can also be framed as part of broader editorial practices within publishing, emphasizing both theoretical grounding and applied assessment. This protocol necessitates the involvement of human judges to validate the texts generated, a process that is often resource-intensive in terms of both time and financial investment, primarily due to the specialized credentials and expertise required of these evaluators. In this paper, we propose an alternative approach by employing LLMs themselves as evaluators within the GrAImes framework. We apply this methodology to assess human-written and AI-generated microfictions in Spanish, to five PhD professors in literature and sixteen literary enthusiasts, and to short stories in both Spanish and English. By comparing the evaluations performed by LLMs with those of human judges, we examine the degree of alignment and divergence between both perspectives, thereby assessing the feasibility of LLMs as auxiliary literary evaluators. Our analysis focuses on the alignment of responses from LLMs with those of human evaluators, providing insights into the potential of LLMs in literary assessment. The conducted experiments reveal that while LLMs cannot be regarded as substitutes for human judges in the evaluation of literary microfictions and short stories, with a Krippendorff'a alpha reliability coefficient less than 0.66, they can serve as a valuable tool that offers an initial perspective on the editorial quality of the texts in question. Overall, this study contributes to the ongoing discourse on the role of artificial intelligence in literature, underlining both its methodological constraints and its potential as a complementary resource for literary evaluation.}
}

	
	
\begin{document}
		
		
		
		\section{Introduction}
	{			
The automatic generation of literary texts is predominantly achieved through the utilization of Large Language Models (LLMs), which are trained on a diverse corpus that includes literary works such as novels, short stories, and~essays alongside general resources like web pages, news articles, and~other textual materials. 
Significant advancements in deep learning and transformer-based architectures have enabled the production of coherent literary texts. As~noted in recent studies \citep{alhussain2021automatic}, these developments have prompted the establishment of both manual and automated validation protocols designed to assess the literary quality of texts generated by artificial intelligence. Such protocols facilitate a comprehensive evaluation of AI-generated literature, allowing for a nuanced understanding of its artistic merit and potential contributions to the literary~landscape.


The gold standard for evaluating literary texts has traditionally been qualified human evaluators, whose expertise is essential for assessing the nuanced qualities of literature. However, this reliance on human judgment is often costly in terms of both time and financial resources, primarily due to the specialized qualifications required of evaluators and the limited availability of such experts. In~light of these constraints, we propose the use of Large Language Models as tools for evaluating literary microfictions authored by both humans and artificial intelligence. Therefore, a~main methodological contribution of this paper is to provide an answer to the following question: to what extent could we rely on LLM as judges in order to evaluate the literary quality of microfictions using existing evaluation protocols such as~GrAImes~\citep{aleman2025can} and TTCW~\citep{chakrabarty2024art}?


Our intention is not to eliminate human judgment from the evaluation process, but~rather to introduce an alternative automated method of assessment that can serve as a preliminary approach to literary curation. This approach seeks to complement rather than replace the~interpretive and aesthetic dimensions that human readers bring to literary analysis. This dual approach allows for a more comprehensive evaluation framework in which LLMs can provide initial insights that complement human expertise. By~embedding LLM-based evaluation within the broader editorial and critical tradition, the~process becomes both scalable and theoretically informed. By~integrating LLMs into the literary evaluation process, we aim to expand the possibilities for literary analysis, making it more accessible and efficient while still honoring the depth and complexity inherent in literary~works.

In our study, we utilize microfictions authored by humans alongside those generated by artificial intelligence, all of which have undergone evaluation by human judges. We then input these texts into LLMs using a prompt that incorporates the GrAImes protocol and specific evaluation instructions. The prompt was carefully designed to mirror the structure and criteria employed by expert human judges, allowing for fair and meaningful comparison between both evaluators. By~comparing the grades assigned by both human evaluators and LLMs, we analyze the validity of LLMs as literary judges, assessing their potential to contribute meaningfully to literary evaluation aesthetic and~to the creative quality of AI-generated fiction. The~analysis also explores the interpretive tendencies and biases exhibited by LLMs when confronted with literary nuance, style, and~affective tone, providing a deeper understanding of their evaluative~behavior.





        
	The rest of this article is organized as follows: Section~\ref{sec:related} outlines the current advancements in text evaluation with large language models as judges; Section~\ref{sec:methods} presents evaluation protocols, datasets, statistical measures, and~evaluator choices used to compare LLM-as-judges with human evaluations; Section~\ref{sec:results} presents the results; finally,~Sections~\ref{sec:discussion} and~\ref{sec:conclusion} discuss the methodological meaning of these results in the task of evaluating AI-generated microfiction and provide insights about further work. }

\section{Related~Work}\label{sec:related}

{The growing body of research on Large Language Models (LLMs) increasingly intersects with questions of creativity, authorship, and~evaluative reliability. In~the context of literary and educational assessment, understanding how LLMs judge, interpret, and~generate text is crucial for determining their legitimacy as evaluators. The~studies reviewed in this section are organized into two conceptual domains---LLMs as judges and LLM benchmarks and quality assessment---each contributing to the broader inquiry into whether machine-based evaluation can approximate human critical judgment.}

\subsection{LLMs as~Judges}

{In addition to creative production, an emerging line of work positions LLMs as evaluators or “judges” of textual quality. Zheng et al.~\citep{zheng2023judging} proposed benchmarks to assess the agreement between LLM-based judgments and human preferences, advancing a methodology for measuring evaluation reliability, while Verga et al.~\citep{verga2024replacing} suggested replacing a single model with a panel of diverse LLMs to reduce bias, conceptualizing a “jury of models” that aggregates multiple perspectives. This approach reflects a shift from individual to collective intelligence in machine evaluation, aligning with traditional peer review structures. Liu et al.~\citep{liu2023g} introduced G-EVAL, a framework that integrates chain-of-thought prompting and human alignment metrics to enhance the interpretability and fairness of model-generated evaluations. Collectively, these works redefine judgment in computational terms, presenting LLMs as scalable evaluators with decisions that can be audited and calibrated against human consensus.}

{Within this context, studies such as Huang~et~al.~\cite{huang2023chatgpt} and Wang~et~al.~\cite{wang2023large} have examined how models assess textual quality, consistency, and~sentiment when compared with human annotators, revealing both promising alignments and persistent biases. Recent frameworks such as the EvaluLLM~\cite{chiang2023can} and LLM-as-a-Judge~\cite{zheng2023judging} benchmarks further formalize this evaluative role, testing how model judgments correlate with human preferences in domains ranging from ethical reasoning to literary style. Chakrabarty~et~al. \citep{chakrabarty2024art} posited a distinct challenge to the narrative of AI creativity by proposing the Torrance Test of Creative Writing (TTCW)~\citep{chakrabarty2024art} as a way to critically evaluate AI-generated stories. By~establishing a rigorous testing standard, this study seeks to clarify the boundaries between genuine creativity and algorithmic output. The~TTCW~\citep{chakrabarty2024art} framework is adapted from human creativity research, thereby bridging psychological and computational perspectives on literary generation.}

{To address biases and enhance LLM evaluations, different papers have proposed innovative methodologies. Liu~et~al.~\cite{liu2023g} introduced a framework that leverages chain-of-thought prompting. This method encourages LLMs to reflect on their reasoning processes, theoretically leading to more aligned evaluations with human judgment. By~explicitly articulating intermediate reasoning steps, these models are expected to produce more transparent and justifiable evaluative outcomes. Verga~et~al.~\cite{verga2024replacing} suggested employing a panel approach in evaluations. This framework aims to mitigate individual biases by aggregating insights from multiple models, achieving a more reliable and balanced assessment of outputs.}

{Further developments in evaluating creative outputs were presented by Chhun~et~al.~\cite{chhun2024language}. Their research investigated the potential for LLMs to self-evaluate and correlate model ratings of their outputs with human annotations, providing insights into the reliability of algorithmic assessments. This line of inquiry brings the discussion full circle, positioning LLMs not only as creators but also as reflexive agents capable of judging their own work in an emerging paradigm where generation and evaluation coexist within the same computational system. Such research opens new philosophical and methodological questions about self-assessment, bias, and~meta-creativity in artificial systems, inviting renewed reflection on what it means for a machine to “appreciate” literature.}

{The convergence of these two domains of generation and evaluation suggests that LLMs may eventually participate in a full literary loop of writing, critiquing, and~revising texts autonomously or in collaboration with humans. This integrated vision positions LLMs not merely as tools but as participants in the evolving ecology of literary production and criticism.}

\subsection{LLM Benchmarks and Quality~Assessment}

{Evaluating the evaluators themselves has become a crucial challenge. Benchmark frameworks such as Chatbot Arena and MT-Bench \citep{zheng2023judging} enable systematic comparison of model performance across subjective and objective tasks, revealing how well LLMs replicate human quality judgments. The~growing use of human preference datasets and meta-evaluation tools ensures that machine assessments remain transparent and reproducible. Assessing the quality of LLM-generated text is not merely a technical requirement but an epistemological one, as it determines the extent to which machine evaluation can be trusted as a proxy for human literary critique.}





{The benchmarks that define the ``LLM-as-a-Judge'' paradigm were critically analyzed by Zheng et al.~\cite{zheng2023judging}. Their research scrutinized the alignment between LLM evaluations and human preferences, aiming to determine the validity and effectiveness of LLMs as judges. Zheng et al. situated the debate within the broader question of whether automated evaluators can meaningfully approximate human aesthetic and ethical reasoning. A fundamental question within this discourse was addressed by Chiang et al.~\cite{chiang2023can}, who developed the EvaluLLM tool. This tool facilitates a customizable evaluation environment that remains under human supervision, aiming to enhance the reliability of automated assessments. Lastly, Pan et al.~\cite{pan2024human} transitioned the focus towards the usability of LLMs in evaluative roles. Their study proposed design principles aimed at fostering productive human--AI collaboration and ensuring that tools designed for judgment can be effectively integrated into academic and professional workflows. Collectively, these frameworks point toward a hybrid evaluative paradigm in which algorithmic efficiency coexists with human interpretive depth.}

{Overall, the~literature reflects a paradigm shift from generation to evaluation, moving from the question of “can AI write?” to the more critical one of “can AI judge writing?”~\cite{li2024leveraging, li2025generation, qin2022survey}. This shift underpins the motivation of the present study, which empirically tests how LLM-based judgments correlate with those of human experts in literary evaluation.}
\section{Concepts, Methods and Materials~Used}\label{sec:methods}
This section outlines the conceptual framework, methodological design, and~materials employed in our study, which aims to investigate the alignment between human and AI-based literary evaluation. Specifically, we evaluate both human-written and AI-generated narratives using two established protocols: GrAImes and the Torrance Test of Creative Writing (TTCW)~\citep{chakrabarty2024art}. These complementary frameworks not only allow us to assess the literary and aesthetic quality of the texts but also to compare how Large Language Models correlate with human judges when acting as evaluators. Our central research question asks: How do LLM-as-judges correlate with human judges in the evaluation of creative writing? 
 To address this, we designed two experiments, one based on microfiction analysis with GrAImes and another on short stories evaluated through TTCW~\citep{chakrabarty2024art}, each testing the extent to which LLM judgments reproduce, diverge from, or~enrich human literary~assessment.

\subsection{Evaluation~Protocols}                        

\subsubsection{GrAImes}

The primary objective of the Grading AI and Human Microfiction Evaluation System (GrAImes)~\citep{aleman2025can} is to integrate literary criteria into the assessment of microfictions generated by both artificial intelligence and human authors. The~GrAImes protocol aims to evaluate the literary quality of AI-generated microfictions. The~concept of reception~\citep{iser1979act, ingarden1993concretizacion} mirrors the editorial processes used by publishers to accept or reject submitted works. GrAImes specifically targets microfictions due to their brevity, which facilitates a more streamlined evaluation~process. 

\textls[-15]{The protocol is organized into three distinct dimensions (see Table~\ref{table:GraAImes_questions} in Appendix 
~\ref{app3})}, each of which is designed to address specific criteria that facilitate a systematic and comprehensive analysis of the texts assigned to evaluators. The~first dimension, referred to as ``story overview and text complexity”, serves to appraise the literary quality of the microfiction by rigorously evaluating several key aspects, including thematic coherence, textual clarity, interpretive depth, and~aesthetic merit. This dimension employs a dual approach that integrates both quantitative metrics such as scoring scales that provide numerical representations of quality and~qualitative assessments encompassing detailed textual commentary that can offer insights into the nuances of literary~value.

The second dimension, known as ``technical evaluation”, focuses on the technical elements that contribute to the overall effectiveness of the narrative. This includes an assessment of linguistic proficiency, which examines the author's command of language; narrative plausibility, which evaluates the believability of the story; and stylistic execution, which considers the author's ability to employ language creatively and effectively. Additionally, this dimension assesses adherence to genre-specific conventions, ensuring that the microfiction aligns with established norms and expectations within its genre, as~well as the effective use of language to convey meaning and evoke emotional responses from the~reader.

The final dimension, named ``editorial/commercial quality”, delves into the commercial viability and editorial appropriateness of the microfiction. This dimension investigates critical factors such as audience appeal, which gauges the potential interest of target readers; market relevance, which assesses the alignment of the work with current trends and demands in the literary marketplace; and the feasibility of publication or dissemination, which considers practical aspects such as distribution channels and potential~readership.

GrAImes~\cite{aleman2025can} distinguishes itself from previous research by implementing a comprehensive evaluation protocol consisting of fifteen targeted questions specifically designed for assessing the literary qualities of microfictions (see Table 
~\ref{table:GraAImes_questions}). The~objective of the present work is to enhance the GrAImes evaluation method, and eventually other methods,
like TTCW~\cite{chakrabarty2024art}, by using LLMs as evaluators through specific evaluation prompts in~addition to literary experts or enthusiasts. This dual application provides a shared evaluative
ground through which human and machine assessments can be systematically compared.
By comparing the results obtained from these diverse evaluators, we aim to provide some
insight into the strengths and limitations of both human and machine-generated fiction.
Such a comparison also allows us to identify dimensions of creativity, coherence, and~stylistic
sensitivity that may diverge between human and model~judgments.

In addition to focusing on microfiction, we broaden our scope by working with
longer short stories. This extension enables us to test the GrAImes framework beyond the
concise structure of microfiction, assessing whether its criteria remain valid for longer and
more complex narratives. This shift allows us to utilize texts previously analyzed in the
Torrance Test of Creative Writing (TTCW) \cite{chakrabarty2023creativity} experiment, enabling a comparative analysis
with another established literary evaluation protocol. By~juxtaposing our findings with
those derived from the TTCW~\cite{chakrabarty2024art}, we seek to expand the discourse surrounding literary
evaluation and the application of LLMs in this domain. Ultimately, the~integration of
both GrAImes and TTCW~\cite{chakrabarty2024art} benchmarks fosters a more holistic understanding of literary
assessment in which computational and humanistic methodologies converge to illuminate
different facets of creativity. This multifaceted approach not only enhances the robustness of
our research but also contributes to a deeper understanding of how LLMs can be effectively
integrated into literary assessment frameworks, ultimately advancing the field of literary
studies and artificial~intelligence.

\subsubsection{TTCW}

The Torrance Test of Creative Writing (TTCW) evaluation protocol for creative writing \citep{chakrabarty2024art} is structured around four pivotal dimensions: fluency, flexibility, originality, and~elaboration. Each dimension encompasses specific criteria that facilitate a comprehensive assessment of narrative quality. Fluency evaluates narrative pacing, understandability, coherence, language proficiency, and~the effective use of literary devices, ensuring that the text flows smoothly and engages the reader. It also considers the narrative ending, assessing its effectiveness in providing resolution and impact, as~well as the balance between scene and summary, which enhances the storytelling experience. Flexibility, on~the other hand, examines the author's adaptability in structural, perspective, and~emotional dimensions, evaluating how well the narrative structure can shift and how effectively the author can evoke a range of emotions through varying perspectives and~voices.

Originality is a critical dimension that assesses the uniqueness of the narrative in terms of theme, thought, and~form. It evaluates how fresh and innovative the ideas presented are as~well as the author's ability to challenge conventional thinking through original insights. Elaboration focuses on the depth of world-building and setting, the~complexity of rhetorical strategies, and~the intricacies of character development. This dimension ensures that the narrative is not only richly detailed but also compelling in its character portrayals and thematic explorations. Together, these dimensions provide a robust framework for a nuanced evaluation of creative writing, fostering a deeper understanding of the artistic and practical qualities that contribute to the overall effectiveness of a~narrative.

  \subsection{Experiments}

  Table~\ref{table:summary} shows the three main experiments we conducted in order to compare human vs. LLM evaluations of fiction's literary and creative quality. Table~\ref{table:stability} shows experiments intended to measure within-model stability, where multiple runs per model with the same prompts were~implemented. 

 
\begin{table}[H]
  \small
  \caption{Summary
  of experiments comparing human and LLM judges using the GrAImes and TTCW~methods.}\label{table:summary}
  
  \begin{tabular}{>{\raggedright\arraybackslash}m{4.1cm} >{\centering\arraybackslash}m{4.2cm} >{\centering\arraybackslash}m{4.3cm}}
    \toprule
    \multicolumn{3}{c}{I. GrAImes Expert Evaluation: Human Experts vs. LLM Judges 
} \\
    \midrule
    Dataset & Human Evaluators & LLM Evaluators \\
    \midrule
     {Six writers'~microfictions Two~expert \newline Two~medium \newline Two~amateur} & Five experts with PhDs in~literature & Five ``expert'' LLMs \\
%
%
    \midrule
    \multicolumn{3}{c}{I. GrAImes Enthusiast Evaluation: Human Enthusiasts vs. LLM Judges}  
    \\
    \midrule
    Dataset & Human Evaluators & LLM Evaluators \\
    \midrule
   {Six AI-generated microfictions \newline Three~by~ChatGPT \newline Three~by~Monterroso} & Sixteen literature enthusiasts & Sixteen ``enthusiast'' LLM prompts \\
%
%
%
\bottomrule
\end{tabular} 
\end{table}

\begin{table}[H]\ContinuedFloat
\small
\caption{\textit{Cont.}}
  \begin{tabular}{>{\raggedright\arraybackslash}m{4.1cm} >{\centering\arraybackslash}m{4.2cm} >{\centering\arraybackslash}m{4.3cm}}
    \toprule
    \multicolumn{3}{c}{III. TTCW and GrAImes protocol comparison} \\   
    \midrule
    Dataset  & Human Evaluators & LLM Evaluators \\
    \midrule
    {Twelve~stories~by TNY \newline Twelve~by~ChatGPT \newline Twelve~by~GPT4 \newline Twelve~by~Claude} & Ten human experts & Four ``expert'' LLMs \\
    \bottomrule
  \end{tabular}
  
\end{table}
\vspace{-9pt}

\begin{table}[H]
  \small
  \caption{Summary 
 of experiments testing within-LLM~stability.}\label{table:stability}
  
  \begin{tabular}{>{\raggedright\arraybackslash}m{3cm} >
  {\raggedright\arraybackslash}m{7.5cm} >
  {\raggedright\arraybackslash}m{2.1cm} }
    \toprule
    \multicolumn{3}{c}{i. GrAImes Expert Evaluation: LLMs-as-Judges Stability} \\
    \midrule
    Dataset & LLM Evaluators & Experiments \\
    \midrule
        Six writers' microfictions Two~expert Two~medium Two~amateur & Claude-sonnet-4-5-20250929,
Gemini-2.5-pro,
o3-2025-04-16,
GPT-4.1-2025-04-14,
Deepseek-v3.2-exp-thinking
 & Five runs as ``expert'' LLM prompts\\
%
%
%
    \midrule
    \multicolumn{3}{c}{ii. GrAImes Enthusiast Evaluation: LLMs-as-Judges Stability} \\
    \midrule
    Dataset & LLM Evaluators& Experiments \\
    \midrule
    \multirow{4}{8em}{Six AI-generated microfictions \newline Three~by~ChatGPT \newline Three~by~Monterroso} &
    Gemini-2.5-pro,
Chatgpt-4o-latest-20250326,
Claude-opus-4-20250514,
Grok-4-0709,
Gemini-2.5-flash,
Llama-4-maverick-17b-128e-instruct,
GPT-4.1-2025-04-14,
o3-2025-04-16,
GPT-5-high,
Deepseek-v3-0324,
Kimi-k2-thinking,
Claude-sonnet-4-20250514-thinking-32k,
Mistral-medium-2508,
GLM-4.6,
QWEN3-235b-a22b-no-thinking,
Mistral-medium-2505
& Five runs as ``enthusiast'' LLM prompts \\
%
%
%
%
    \midrule
    \multicolumn{3}{c}{iii. TTCW and GrAImes Protocol Comparison: LMMs-as-Judges Stability} \\       
    \midrule
    Dataset  & LLM Evaluators & Experiments \\
    \midrule
    Twelve~stories~by TNY Twelve~by~ChatGPT Twelve~by~GPT4 Twelve~by~Claude & Claude-opus-4-1-20250805,
ChatGPT-4o-latest-20250326, 
GPT-5-high & Five runs as ``expert'' LLM prompts \\

    \bottomrule
  \end{tabular}
  
\end{table}

  \subsubsection{GrAImes Expert Evaluation: Human Experts vs. LLM~Judges}

The original GrAImes dataset~\citep{aleman2025can} contains evaluations by five literary experts holding PhDs in literature and working on an academic permanent position who responded to the standard GrAImes evaluation questionnaire. These evaluators assessed six human-written short~stories. 

We chose five LLMs (see Table~\ref{table:llms_exps_lmarena}) and gave them an expert prompt (see Section~\ref{sec:datasets}) in order to generate LLM-as-judge assessments. We analyze and compare the answers in the tables and graphics shown in Section~\ref{sec:results}.  This cross-evaluation allowed us to quantify the correlation between human and model-based scoring, observing not only numerical agreement but also convergence in qualitative~justification.

  \subsubsection{GrAImes Enthusiast Evaluation: Human Enthusiasts vs. LLM~Judges}
  On the original GrAImes experiments, sixteen human evaluators who were members of a literature enthusiasts YouTube channel~\citep{aleman2025can} answered the GrAImes questionnaire on six AI-generated microfictions. Their qualitative responses were compared between the experts and the empirical readers. These comparative results provided a reference distribution of human judgments that later served as a baseline for LLM evaluation under identical~conditions.

    We then gave ``enthusiast'' prompts (see Section~\ref{sec:datasets}) to sixteen different LLMs (Table~\ref{table:llms_enths_lmarena}) to analyze and compare their answers with the human~ones.

\subsubsection{TTCW and GrAImes Protocol~Comparison}

We conducted another experiment to compare GrAImes with the Torrance Test for Creative Writing (TTCW) ~\citep{chakrabarty2024art}. The original TTCW~\citep{chakrabarty2024art} dataset includes forty-eight short stories (12 written by professionals selected from The New Yorker collection and 36~generated with three different LLMs); see Section~\ref{sec:datasets}.

To facilitate a direct comparison between the GrAImes and TTCW~\citep{chakrabarty2024art} evaluation protocols, we used the same short stories for both methods. However, we translated the GrAImes questions and prompts into English to align with the TTCW~\citep{chakrabarty2024art} format. GrAImes employs a Likert scale ranging from 1 (strongly disagree) to 5 (strongly agree), while the TTCW~\citep{chakrabarty2024art} uses a binary yes/no response format accompanied by a written justification. To~align the evaluation grades, we converted the Likert scale responses into binary answers: grades of 1 and 2 were assigned a value of 0 (no), grades of 4 and 5 were assigned a value of 1 (yes), and~grades of 3 were distributed according to the average derived from the newly-assigned 0 and~1 scores.

Applying both GrAImes and TTCW~\citep{chakrabarty2024art} allows  the editorial--literary and creativity--psychological evaluative paradigms to be comared, thereby deepening our understanding of how LLMs interpret literary merit under different theoretical~lenses.

\subsubsection{LLMs Within-Models~Stability}
To ensure the stability and reliability of the LLMs deployed in our research, we conducted a series of repetitions; specifically, we executed each LLM five times using identical prompts and microfictions to apply the GrAImes evaluation protocol, emulating the conditions presented to the LLMs in each of the tests that we are presenting. This systematic repetition allowed us to assess the consistency of the models' outputs across multiple~iterations.

{
\subsubsection{Inter-Rater~Reliability}

    In the GrAImes presentation article \citep{aleman2025can}, the inter-rater reliability Kendall’s W analysis of GrAImes sections answers provided by literary experts revealed disparate levels of consensus.~The highest degree of concordance was evident for MF1 and MF2, both of which were authored by the expert writer; conversely, MF3 and MF5, which were penned by less seasoned authors, exhibited lower levels of agreement. These results imply a correlation between author expertise and the assessments provided by literary experts. These findings support existing research \citep{mccutchen2011novice} showing that higher writing expertise correlates with better-structured and logically consistent texts, while lower expertise results in fragmented writing. Judges rated microfictions by more experienced authors higher than those by emerging authors, aligning with the goal of our evaluation protocol to quantitatively and qualitatively assess microfictions based on their literary~merit. 
    
    The ICC analysis of GrAImes questions responses in \citep{aleman2025can} showed varying reliability among sixteen literature enthusiast raters evaluating AI generated texts. Three questions had poor reliability (ICC < 0.50), with~Question 8 showing severe inconsistency (ICC -0.44). Questions 5 and 6 had excellent reliability (ICC > 0.90), while Questions 9, 11, 12, and~13 had moderate reliability (ICC 0.6 –0.70). }


\subsection{Datasets}
\label{sec:datasets}
For the GrAImes expert evaluation, 
we applied the six Spanish microfictions used in \citep{aleman2025can} along with the set of fifteen questions from the GraImes protocol. These microfictions were written by three different people: an author with published books (MF1 
and 2), an~author with publications in anthologies and literary magazines (MF 3 and 4), and~an emerging writer (MF 5 and 6). These six microfictions were fed to the same number of LLMs along with GrAImes in a prompt with instructions to evaluate them as an expert; this was repeated three times for every LLM according to the three sections of the GrAImes evaluation protocol, each section specified with its belonging questions and their corresponding Likert scale or open answer. This design ensured that both human and machine evaluators followed identical procedural conditions, strengthening the validity of our comparative~analysis.
  
For the TTCW~\citep{chakrabarty2024art}-GrAImes comparison, we used the data available from the TTCW~\citep{chakrabarty2024art} experiments, in which the authors curated a collection of twelve short stories from The New Yorker spanning the period from 2020 to 2023. The~selection criteria were designed to encompass a diverse array of authors and narrative plots, ensuring wide representation of contemporary literary voices. A~one-sentence plot summary was also generated by GPT-4 according to the selected short stories. Several LLMs (GPT-3.5, GPT-4, and~Claude V1.3) were prompted to generate twelve~original stories each that mirrored the length of each selected New Yorker narrative, utilizing the provided one-sentence plot summaries as a foundation. We then evaluated these 48 short stories (12 human-written, 36 AI-generated) using the TTCW~\citep{chakrabarty2024art} framework to examine how creativity dimensions align or diverge across~evaluators.

\subsubsection*{Large Language Models as~Judges 
}

We selected five LLMs for the GrAImes Expert evaluation (Experiment I) and  sixteen LLMs (see Table~\ref{table:llms_exps_lmarena}) for the GrAImes Enthusiast Evaluation (Experiment II). The~LLM selection criterion was their ranking in the leaderboard of the Creative Writing category from LMArena ({\url{https://lmarena.ai/}}
)~\citep{chiang2024chatbot}, which is a web-based platform that evaluates LLMs while allowing human-only interaction. Five LLMs were used for evaluation as literary experts using GrAImes with human-written microfictions (see Table~\ref{table:llms_enths_lmarena}), and sixteen LLMs were prompted to simulate a literature enthusiast evaluating AI-generated microfictions. We used an expert's prompt for each question from GrAImes;~later, we compared them with those of the five human experts, an~enthusiast's prompt to compare with the human enthusiasts, and a prompt specifically designed for the short stories from TTCW~\cite{chakrabarty2024art} used with GrAImes and the LLMs listed in Table~\ref{table:llms_ttcw_graimes_lmarena}. These prompts can be reviewed in Appendix \ref{Verbatim_prompts}.

\subsection{{Limitations}}

{This computational framework faces inherent limitations stemming from the epistemological constraints of LLMs. Critical issues include linguistic/cultural diversity and the risk of perpetuating training data biases. LLMs' potential role as computational literary gatekeepers demands careful scrutiny, particularly regarding training data scope, genre adaptability, and~semantic drift in generative~models.

The efficacy of LLMs is constrained by several factors. Data size limitations cap the text length that it can process in one interaction, affecting its handling of lengthy or complex inputs. There is also a concern about a future shortage of high-quality human-generated data, which could impede the model's improvement. Genre scope limitations cause struggles with specialized or underrepresented genres, leading to suboptimal performance in niche writing styles. \textls[-15]{Additionally, contextual drift can degrade the model's coherence and relevance over extended conversations, resulting in less accurate~outputs}.}

{The number of literary experts (5) and literature enthusiasts (16) in the paper that introduced GrAImes~\cite{aleman2025can} was due to the required specialty and the difficulty of engaging PhD literary professors, as~well as to the scarcity of frequent readers available to answer the questionnaires and~the lack of monetary resources assigned to this. Due to the use of lmarena.ai as the platform where the LLMs were available, it was not possible to modify the LLM model parameters (temperature, max tokens, top-p, seed).
}

{During the preparation of this work, the authors used the Mistral Small 3 open source language model provided by Duck.ai
  and~the Language Tool
   provided by Mozilla to improve the style of some sentences. The authors also used the large language models claude-sonnet-4-5-20250929, Gemini-2.5-Flash-Preview-04-17, o3-2025-04-16, DeepSeek-V3-0324, GPT-4.1-2025-04-14, gemini-2.5-pro, chatgpt-4o-latest-20250326, claude-opus-4-20250514, grok-4-0709, gemini-2.5-flash, gemini-2.5-flash-lite-preview-06-17-thinking, gpt-4.1-2025-04-14, o3-2025-04-16, grok-3-preview-02-24, deepseek-v3-0324, deepseek-r1-0528, claude-sonnet-4-20250514-thinking-32k, kimi-k2-0711-preview, hunyuan-turbos-20250416, qwen3-235b-a22b-no-thinking, mistral-medium-2505, claude-opus-4-1-20250805, chatgpt-4o-latest-20250326, and gpt-5-high for the purposes of the evaluation of microfictions and short stories applying GrAImes and to obtain the results used in this study. After~using these tools, the~authors reviewed and edited the content as needed, and take full responsibility for the content of the publication.}
            
\section{Results}
\label{sec:results}

This section presents the results of our experiments, which aimed to compare human and LLM-based evaluations of literary texts under two complementary frameworks: GrAImes and the Torrance Test of Creative Writing (TTCW)~\citep{chakrabarty2024art}. We analyze how literary scholars and language models assessed both human-written and AI-generated stories, focusing on three main aspects: (1) the degree of alignment between human and model ratings, (2) the qualitative tendencies observed in narrative interpretation and technical assessment, and~(3) the extent to which LLMs can approximate expert-level literary judgment. The~results are organized by experimental phase, beginning with the GrAImes evaluation of human-written microfictions and continuing with the analysis of model-based evaluations under identical~conditions.

\subsection{GrAImes Expert Evaluation Results (Experiment I)}


Based on the responses obtained from the five literary experts who used GrAImes with six Spanish microfictions and~presented in \citep{aleman2025can}, we conclude that literary experts assigned higher ratings to those microfictions (1 and 2) authored by published writers. However, the~data revealed a high standard deviation, indicating that while the evaluations were predominantly favorable, there was considerable variability among the experts' assessments. In~contrast, the~lowest-ranked microfictions (3 and 6) which exhibited a lower average response score also demonstrated a reduced standard deviation, suggesting greater consensus among the judges regarding these texts. Notably, these lower-ranked works were authored by writers published in literary magazines or by small-scale editorial presses with limited print runs. These~findings imply a direct correlation between author expertise and the the texts' internal consistency, highlighting the influence of established literary credentials on evaluative agreement among~experts.

These observations served as the benchmark for subsequent comparison with LLM-based evaluations, providing a human reference against which the machine judgments were interpreted. In~particular, distribution of the~human experts’ scores offers insights into the expected variability and depth of literary judgment, allowing us to contextualize LLM performance beyond mere numerical~correlation.


The analysis provided by the LLMs in response to human-written microfictions revealed intriguing insights into their evaluative capabilities across various dimensions. As~illustrated in \cref{table:chatbots_humanw,table:chatbots_quest_SD_humanw,table:Chatbots_mf_h_sects__sum_AV_SD}, the~responses to the question regarding the proposal of alternative interpretations beyond the literal yielded consistently high average ratings, particularly for microfictions 1 and 2, which received perfect scores of 5.0. This suggests that these texts were perceived as rich in interpretive depth, allowing for multifaceted engagement with the narrative. Conversely, microfictions 3 and 6, while still receiving favorable evaluations, exhibited lower average scores, indicating a potential limitations in their complexity. The~standard deviations across these responses, particularly the low variability for microfictions 1 and 2, further emphasizes consensus among the evaluators regarding the literary merit of these~works.

Concerning the credibility of the stories, the scores varied significantly, with~microfictions 3 and 6 receiving notably lower average scores (2.4 and 2.2, respectively). This disparity highlights the challenges LLMs face in evaluating narrative plausibility, particularly for texts that may not align with conventional storytelling norms. However, the~responses to questions about the texts' ability to evoke new visions of reality and genre indicated a generally positive reception, with~average scores around 4.0. This suggests that despite some inconsistencies in credibility, the~microfictions were still recognized for their innovative contributions to literary discourse. Furthermore, the~editorial and commercial quality metrics revealed a strong inclination among evaluators to recommend and share these texts, with~average scores exceeding 4.0 for most~microfictions.

Overall, the~LLM evaluations reflect a partial but notable convergence with human expert assessments. The~models demonstrated sensitivity to interpretive richness and stylistic quality, yet displayed limitations in assessing narrative plausibility and contextual subtlety. These results suggest that LLMs, while not substitutes for human judgment, can emulate aspects of expert evaluation and provide preliminary insights consistent with established literary critique. In~the following sections, we extend this comparative analysis to AI-generated stories and examine how the correlation between LLM and human evaluations evolves across different~protocols.

\begin{table}[H]
	\small
	\caption{LLM responses to questions human-written~microfictions (Likert scale).}
	\label{table:chatbots_humanw}
\begin{adjustwidth}{-\extralength}{0cm}

	\begin{tabular}{
			m{24.3em} m{1em}m{1em}
			m{1em} m{1em} m{1em} m{1em} 
			m{1em} m{1em} m{1em} m{1em}
			m{1em} m{1em} m{1em} m{1em}
		} 
		\toprule
		\multicolumn{15}{c}{\textbf{LLM responses to human-written microfictions}}\\
		\midrule
		& \multicolumn{2}{c}{\textbf{MF1 
        }} & \multicolumn{2}{c}{\textbf{MF2}}         
        & \multicolumn{2}{c}{\textbf{MF3}}         
        & \multicolumn{2}{c}{\textbf{MF4}}
        & \multicolumn{2}{c}{\textbf{MF5}}
        & \multicolumn{2}{c}{\textbf{MF6}}
        & \multicolumn{2}{c}{\textbf{Avg.}}\\
		\midrule
		\textbf{Question} & \textbf{AV} & \textbf{SD} & \textbf{AV} & \textbf{SD} & \textbf{AV} & \textbf{SD}  & \textbf{AV} & \textbf{SD} & \textbf{AV} & \textbf{SD} & \textbf{AV} & \textbf{SD} & \textbf{AV} & \textbf{SD} \\
		\midrule
		\multicolumn{15}{c}{ Story Overview and Text Complexity} 
\\
		\midrule
		3.-Does it propose other interpretations, in~addition to the literal one? 
		&5.0 & 0.0 & 5.0 & 0.0 & 3.8 & 0.4 & 4.0 & 0.7 & 3.6 & 0.5 & 4.4 & 0.9 & 4.3 & 0.4\\		
		\midrule
		\multicolumn{15}{c}{Technical Assessment} 
 \\
		\midrule
		5.-Is the story credible? 
		&2.4 & 1.1 & 2.6 & 1.1 & 4.0 & 0.7 & 4.0 & 0.7 & 3.0 & 0.7 & 2.2 & 0.8 & 3.0 & 0.9 \\
		\midrule
		6.-Does the text require your participation or cooperation to complete its form and meaning?
		&4.4 & 0.5 & 4.6 & 0.5 & 3.2 & 0.4 & 4.0 & 0.7 & 3.2 & 0.8 & 4.6 & 0.9 & 4.0 & 0.7 \\		 		 
		\midrule
		7.-Does it propose a new vision of reality? 
		&4.4 & 0.5 & 4.6 & 0.5 & 3.0 & 1.0 & 3.8 & 0.4 & 3.6 & 1.1 & 3.6 & 0.5 & 3.8 & 0.7 \\
		\midrule
		8.-Does it propose a new vision of the genre it uses?  
		&4.2 & 0.8 & 4.0 & 0.7 & 2.6 & 0.5 & 3.4 & 0.5 & 3.4 & 0.5 & 4.0 & 1.2 & 3.6 & 0.7 \\		
		\midrule
		9.-Does it propose a new vision of the language itself? 
		&4.8 & 0.4 & 4.8 & 0.4 & 3.6 & 0.5 & 4.4 & 0.5 & 3.2 & 0.4 & 4.0 & 0.7 & 4.1 & 0.5 \\		
		\midrule
		\multicolumn{15}{c}{Editorial/Commercial Quality}  \\	
		\midrule
		10.-Does it remind you of another text or book you have read?
		&4.0 & 0.0 & 4.0 & 0.7 & 3.2 & 1.1 & 3.6 & 0.5 & 4.0 & 1.0 & 3.8 & 0.8 & 3.8 & 0.7 \\ 
		\midrule
		11.-Would you like to read more texts like this? 
		&4.8 & 0.4 & 4.6 & 0.5 & 3.0 & 0.7 & 3.0 & 0.0 & 4.0 & 0.8 & 4.2 & 0.7 & 3.9 & 0.5 \\		
		\midrule 
		12.-Would you recommend it? 
		&4.8 & 0.4 & 4.6 & 0.5 & 3.4 & 0.9 & 4.2 & 0.4 & 4.2 & 0.8 & 4.0 & 0.7 & 4.2 & 0.6 \\		
		\midrule
		13.-Would you give it as a present? 
		&4.4 & 0.5 & 4.2 & 0.8 & 2.6 & 1.1 & 3.8 & 0.8 & 3.8 & 1.1 & 3.4 & 1.1 & 3.7 & 0.9 \\		
		\bottomrule		
	\end{tabular}
\end{adjustwidth}
\end{table}
\vspace{-9pt}

\begin{table}[H]
	\small	
	\caption{LLMs responses to human-written microfictions (by standard deviation, ascending order).}
	\label{table:chatbots_quest_SD_humanw}
\begin{adjustwidth}{-\extralength}{0cm}

	\begin{tabularx}{\fulllength}{lCC}
		\midrule
		\multicolumn{3}{c}{\textbf{LLM Responses to Human-Written Microfictions, Ordered by SD}}\\
		\midrule
		\textbf{Question} 
		& \textbf{AV} & \textbf{SD} \\
		\midrule			
		3.-Does it propose other interpretations, in~addition to the literal one? 
		& 4.3 & 0.4 \\
		\midrule
		11.-Would you like to read more texts like this? 
		& 3.9 & 0.5 \\
		\midrule
		9.-Does it propose a new vision of the language itself? 
		& 4.1 & 0.5 \\
		\midrule
		12.-Would you recommend it? 
		& 4.2 & 0.6 \\
		\midrule
		6.-Does the text require your participation or cooperation to complete its form and meaning? 
		& 4.0 & 0.7 \\
		\midrule
		7.-Does it propose a new vision of reality?  
		& 3.8 & 0.7 \\
		\midrule
		10.-Does it remind you of another text or book you have read? 
		& 3.8 & 0.7 \\
		\midrule 
		8.-Does it propose a new vision of the genre it uses? 
		& 3.6 & 0.7 \\
		\midrule
		13.-Would you give it as a present?  
		& 3.7 & 0.9 \\
		\midrule 
		5.-Is the story credible?
		& 3.0 & 0.9 \\
		\bottomrule 				
	\end{tabularx}
\end{adjustwidth}
\end{table}
\vspace{-9pt}

		\begin{table}[H]
	\caption{LLM 
 GrAImes sections summarized AV and SD for human-written microfictions.}
	\small
	\label{table:Chatbots_mf_h_sects__sum_AV_SD}
\begin{adjustwidth}{-\extralength}{0cm}

	\begin{tabularx}{\fulllength}{CCCCCCCCCCCCCCCC} 
		\toprule
		\multicolumn{4}{c}{\textbf{Story Overview}} & & \multicolumn{3}{c}{\textbf{Technical Assessment}} & &  \multicolumn{3}{c}{\textbf{Editorial/Commercial}} & & \multicolumn{3}{c}{\textbf{Total Analysis}}  \\
		\cmidrule{1-4}\cmidrule{6-8}\cmidrule{10-12}\cmidrule{14-16}
		\#& \textbf{MF} & \textbf{AV} & \textbf{SD} & & \textbf{MF} & \textbf{AV} & \textbf{SD} & & \textbf{MF} & \textbf{AV} & \textbf{SD} & & \textbf{MF} & \textbf{AV} & \textbf{SD}\\ 
		\cmidrule{1-4}\cmidrule{6-8}\cmidrule{10-12}\cmidrule{14-16}
		1& 1 & 5.0 & 0.0 & & 2 & 4.0 & 0.7 & & 1 & 4.5 & 0.4 & & 1 & 4.3 & 0.5 \\
		\cmidrule{1-4}\cmidrule{6-8}\cmidrule{10-12}\cmidrule{14-16}
		2& 2 & 5.0 & 0.0 & & 1 & 4.1 & 0.7 & & 2 & 4.4 & 0.7 & & 2 & 4.3 & 0.6 \\
		\cmidrule{1-4}\cmidrule{6-8}\cmidrule{10-12}\cmidrule{14-16}
		3& 6 & 4.4 & 0.9 & & 4 & 3.9 & 0.6 & & 5 & 4.0 & 0.9 & & 4 & 3.8 & 0.5 \\
		\cmidrule{1-4}\cmidrule{6-8}\cmidrule{10-12}\cmidrule{14-16}
		4& 4 & 4.0 & 0.7 & & 6 & 3.7 & 0.8 & & 6 & 3.9 & 0.8 & & 6 & 3.8 & 0.8 \\
		\cmidrule{1-4}\cmidrule{6-8}\cmidrule{10-12}\cmidrule{14-16}
		5& 3 & 3.8 & 0.4 & & 3 & 3.3 & 0.6 & & 4 & 3.7 & 0.5 & & 5 & 3.6 & 0.8 \\
		\cmidrule{1-4}\cmidrule{6-8}\cmidrule{10-12}\cmidrule{14-16}
		6& 5 & 3.6 & 0.5 & & 5 & 3.3 & 0.7 & & 3 & 3.1 & 1.0 & & 3 & 3.2 & 0.8 \\
		\bottomrule
	\end{tabularx}
\end{adjustwidth}
\end{table}


    The results reveal a distinct connection between author skill level and the internal consistency of the texts. The~microfictions from expert authors (MF1 and MF2)         
    received the highest scores on the Likert scale. This suggests that expert authors create texts that are more coherent and internally consistent, which is consistent with prior studies associating higher authorial expertise with logically connected and well-structured~writing.

    {We used Krippendorff's Alpha ($\alpha$)~\citep{krippendorff1970estimating} to obtain the agreement and 95\% confidence intervals (CIs)  of the LLMs used with GrAImes and human-written microfictions. The~result of 0.016 suggests that the agreement in the responses obtained with the language models is no better than chance.

The analysis of Krippendorff's $\alpha$ coefficients and confidence intervals across the various large language models and five literary experts,~presented in \cref{table:Krippendorf_alpha_le1_lllms_hw_mfs,table:Krippendorf_alpha_le1_2_llms_hw_mfs,table:Krippendorf_alpha_le1_3_llms_hw_mfs,table:Krippendorf_alpha_le1_4_llms_hw_mfs,table:Krippendorf_alpha_le1_5_llms_hw_mfs}, reveals significant insights into the evaluative alignment between machine-generated assessments and human literary judgment}. For~instance, the~performance of the Grok-3-Preview-02-24 model shows considerable variability, with~coefficients ranging from a low of $-$0.520 for microfiction 6 to a high of 0.197 for microfiction 3 when evaluated by literary expert 1. This suggests that while some microfictions resonated with the model's evaluative framework, others did not, indicating a potential disconnect between the model's interpretative capabilities and the nuanced expectations of human evaluators. In~contrast, the~Gemini-2.5-Flash-Preview-04-17 model demonstrated more stable performance, particularly with microfiction 2, which received a Krippendorff's $\alpha$ for agreement with human judges of 0.584{.} 

The analysis of Krippendorff's $\alpha$ coefficients for literary expert 2,~shown in \mbox{Table~\ref{table:Krippendorf_alpha_le1_2_llms_hw_mfs}}, reveals a complex relationship between human literary judgment and the LLMs' evaluations. Notably, the~Gemini-2.5-Flash-Preview-04-17 model achieved a coefficient of 0.208 for microfiction 2, indicating moderate alignment with the expert's assessment. However, the~overall performance of this model across the other microfictions was less consistent, with~several negative coefficients, particularly for microfiction 4 ($-$0.275) and microfiction 3 ($-$0.226). This inconsistency suggests that while some texts resonated with the model's evaluative framework, others did not, highlighting the challenges LLMs face in capturing the nuanced qualities that human evaluators prioritize. The~negative coefficients for Grok-3-Preview-02-24 across multiple microfictions further emphasizes the potential limitations of this model in aligning with human literary standards, particularly in the context of microfictions that may require a deeper interpretative~engagement.

Similarly, the~results for literary expert 3,~presented in Table~\ref{table:Krippendorf_alpha_le1_3_llms_hw_mfs}, further illustrate the variability in LLM performance. The~coefficients for this expert reveal a predominance of negative values, particularly for microfictions 3 ($-$0.196) and 4 ($-$0.230), indicating significant divergence from the expert evaluations. The~highest coefficient recorded for this expert was a modest 0.050 for microfiction 1 when assessed by Grok-3-Preview-02-24, suggesting that even the most favorable evaluations do not reflect strong 
{agreement} with human judgment. This pattern raises critical questions about the capacity of LLMs to effectively evaluate literary texts, particularly those that demand a nuanced understanding of narrative complexity and thematic~depth.

\begin{figure}[H]

   \hspace*{-6pt} \includegraphics[width=0.5\linewidth]{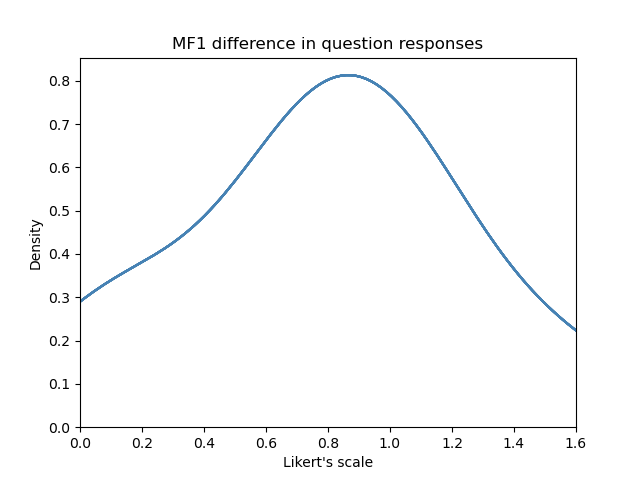}
    \includegraphics[width=0.5\linewidth]{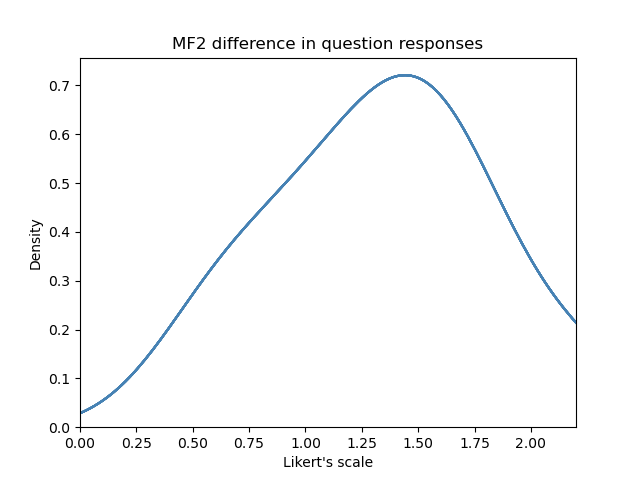}\\
      \hspace*{-6pt} \includegraphics[width=0.5\linewidth]{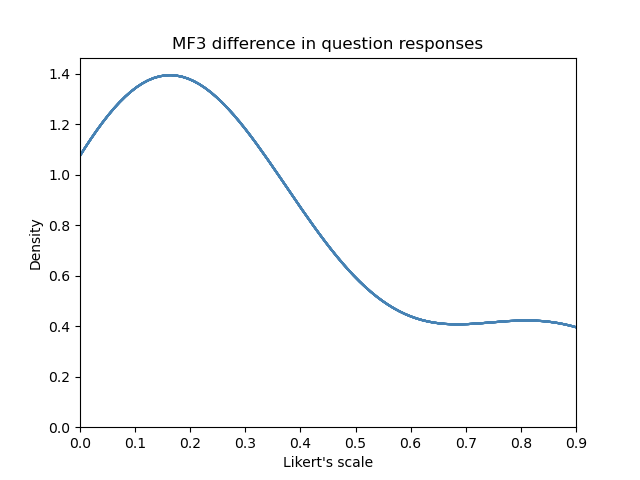}
    \includegraphics[width=0.5\linewidth]{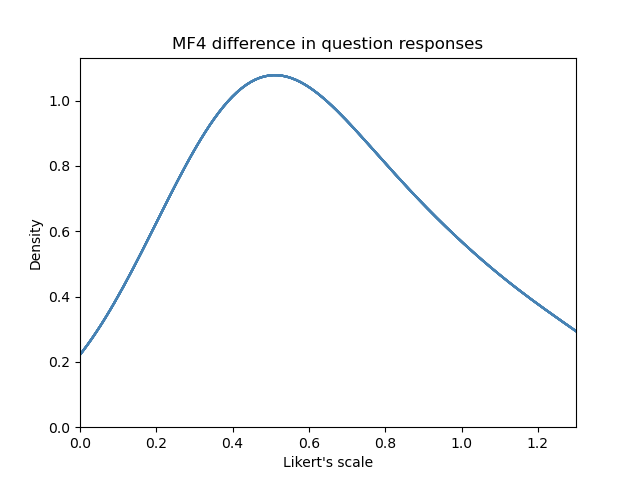}\\
      \hspace*{-6pt} \includegraphics[width=0.5\linewidth]{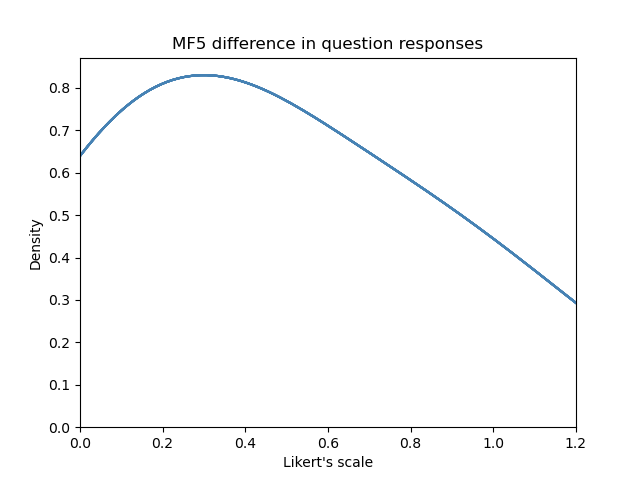}
    \includegraphics[width=0.5\linewidth]{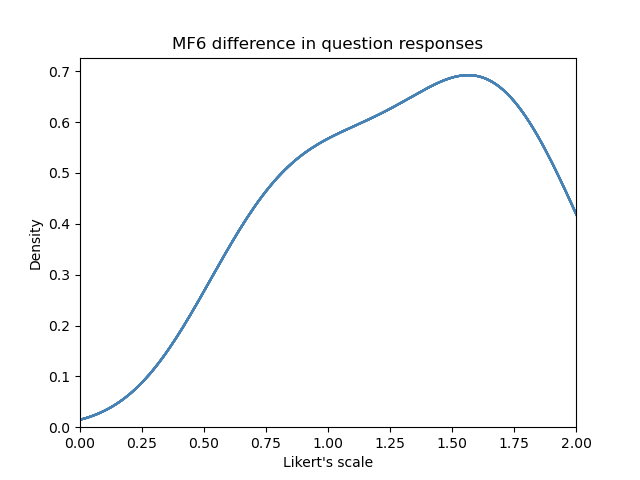}
		    \caption{Differences 
 between human experts and LLMs in AV responses by~microfiction.}
    \label{fig:density_diff_exps}
\end{figure}

The results across the different literary experts further explain the complexities of LLM evaluations. For~example, as shown in Table~\ref{table:Krippendorf_alpha_le1_5_llms_hw_mfs}, literary expert 5 consistently reported negative coefficients for most microfictions when assessed by the various LLMs, indicating a significant divergence in judgment. This pattern raises questions about the inherent biases or limitations of the models in capturing the subtleties of literary quality as perceived by human experts. Conversely, as~presented in Table~\ref{table:Krippendorf_alpha_le1_4_llms_hw_mfs}, literary expert 4 noted a positive coefficient of 0.445 for microfiction 3 when evaluated by Grok-3-Preview-02-24, suggesting that certain texts may align more closely with the model's evaluative criteria. Overall, these findings underscore the lack of alignment between LLMs and human literary standards of quality. This can be confirmed in the line histograms presented in Figure~\ref{fig:density_diff_exps} 
, which display the differences between the literary experts and the LLMs' grades on the Likert scale by applied GrAImes question and microfiction.

\begin{table}[H]
	\caption{Krippendorff's $\alpha$ and CI of LLMs and literary expert 1 using GrAImes with human-written MFs. LLM 1 = Grok-3-Preview-02-24, LLM 2 = Gemini-2.5-Flash-Preview-04-17, 
LLM 3 = o3-2025-04-16, LLM 4 = DeepSeek-V3-0324, 
LLM 5 = GPT-4.1-2025-04-14
.}
	\label{table:Krippendorf_alpha_le1_lllms_hw_mfs}
	\scriptsize

\begin{adjustwidth}{-\extralength}{0cm}

	\begin{tabular}{p{.27cm} p{.5cm} p{1.65cm} 
    p{.5cm} p{1.6cm} 
    p{.5cm} p{1.6cm} 
    p{.5cm} p{1.6cm} 
    p{.5cm} p{1.6cm} 
    p{.5cm} p{1.65cm} }
		\midrule
		\multicolumn{13}{c}{\textbf{Krippendorff's \boldmath{$\alpha$} and Confidence Interval (CI) for Literary Expert 1}}  \\
		\midrule
            \textbf{LLM} & \textbf{MF1}&& \textbf{MF2}&& \textbf{MF3}&&\textbf{MF4}&& \textbf{MF5}&&\textbf{MF6}& \\ 
		\midrule
		 & \boldmath{$\alpha$}&\textbf{CI}& \boldmath{$\alpha$} & \textbf{CI}& \boldmath{$\alpha$} & \textbf{CI}& \boldmath{$\alpha$} & \textbf{CI}& \boldmath{$\alpha$}  & \textbf{CI}& \boldmath{$\alpha$} & \textbf{CI} \\ 
		\midrule		
        1 &0.112 &  ($-$0.028, 0.468) & 0.136 &   ($-$0.027, 0.199)& 0.197 &  ($-$0.022, 0.252)& $-$0.236 &   (0.007, 0.442)& 0.058&  ($-$0.045, 0.206) & $-$0.520 &   ($-$0.038, 0.392)\\ \midrule
        2& 0.010 &   ($-$0.044, 0.168))& 0.584 &  ($-$0.048, 0.135)&  0.158 &   ($-$0.021, 0.268) &$-$0.008 &   ($-$0.030, 0.410) & 0.101 &   ($-$0.013, 0.286) & 0.022 &   ($-$0.001, 0.315)\\ \midrule
        3 & 0.128 &   ($-$0.056, 0.296) & $-$0.118 &   ($-$0.047, 0.243) & $-$0.163 &   (0.052, 0.444) & $-$0.126 &  ($-$0.013, 0.332) &  0.228 &   ($-$0.046, 0.296) & $-$0.291 &  ($-$0.024, 0.454)\\ \midrule
        4&$-$0.288 &  ($-$0.038, 0.678)& $-$0.008 &   (0.007, 0.413) & $-$0.258 &   (0.070, 0.386) & $-$0.008 &   ($-$0.030, 0.410) & $-$0.171 &   (0.039, 0.340) & $-$0.078 &   (0.138, 0.634)\\ \midrule
        5 & 0.188 &   ($-$0.022, 0.620) & 0.168 &   ($-$0.019, 0.353) & $-$0.293 &  (0.075, 0.465) &  $-$0.188 &   ($-$0.019, 0.309) & 0.058 &   ($-$0.045, 0.206) &  0.202 &   ($-$0.044, 0.467))\\ \midrule

	\end{tabular}	
\end{adjustwidth}
\end{table}

\begin{figure}[H]
       \hspace*{-6pt}\includegraphics[width=.5\linewidth]{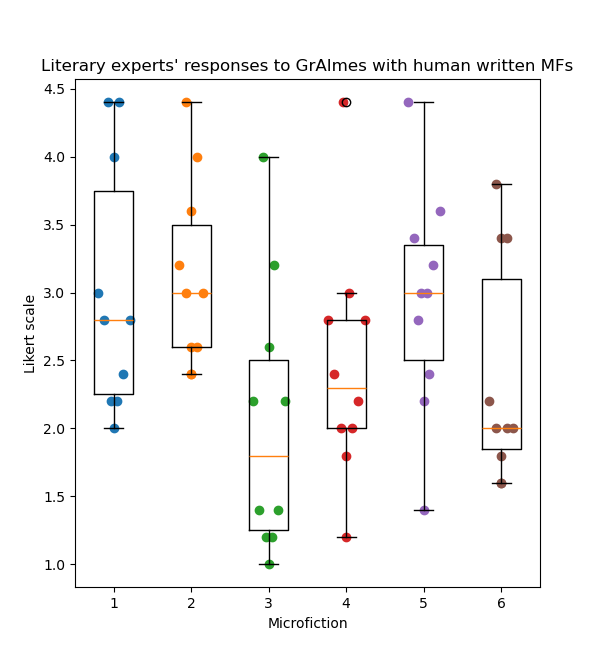}
	\includegraphics[width=.5\linewidth]{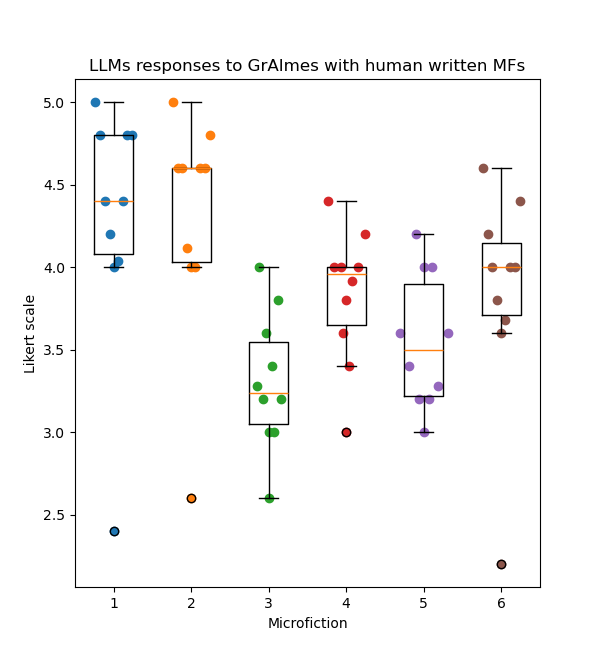}

	\caption{Comparison 
 between human and LLM GrAImes expert assessment of human-written microfictions (Experiment I).}
	\label{fig:LLMs_beeswarm_hw_mfs}
\end{figure}

Figure~\ref{fig:LLMs_beeswarm_hw_mfs}  reveals a consistent and significant evaluative gap between human literary experts and LLMs in their assessment of human-written microfictions. The human experts demonstrated a consistent level of appreciation for the human-authored texts due to the differences in the writers expertise, maintaining a stable and superior average rating of  2.6 across the evaluated samples. Conversely, the~LLMs' responses were not only higher at around 3.8, but~also showed greater variability, as~indicated by the fluctuating orange line. This suggests that while the human experts consistently applied a stable set of criteria valuing artistic merit, the~LLMs' evaluation process was less consistent and more sensitive to variations in textual features that may not align with human notions of literary~quality.

This consistent disparity highlights a fundamental incongruity in the evaluative frameworks employed by humans and AI models. The~LLMs' lower and more volatile scores suggest that their assessment is likely driven by statistical, syntactic, or~semantic pattern-matching against their training data rather than by an appreciation for the narrative cohesion, emotional resonance, and~creative innovation recognized by experts. The~stability of the expert ratings underscores a consensus on what constitutes quality microfiction, a~consensus that current LLMs fail to replicate. This indicates that LLMs' ability to generate human-like text does not equate to an ability to evaluate it with human-like~judgment.

\begin{figure}[H]
   \vspace{-3pt}\hspace{-30pt}
\includegraphics[width=1\linewidth]{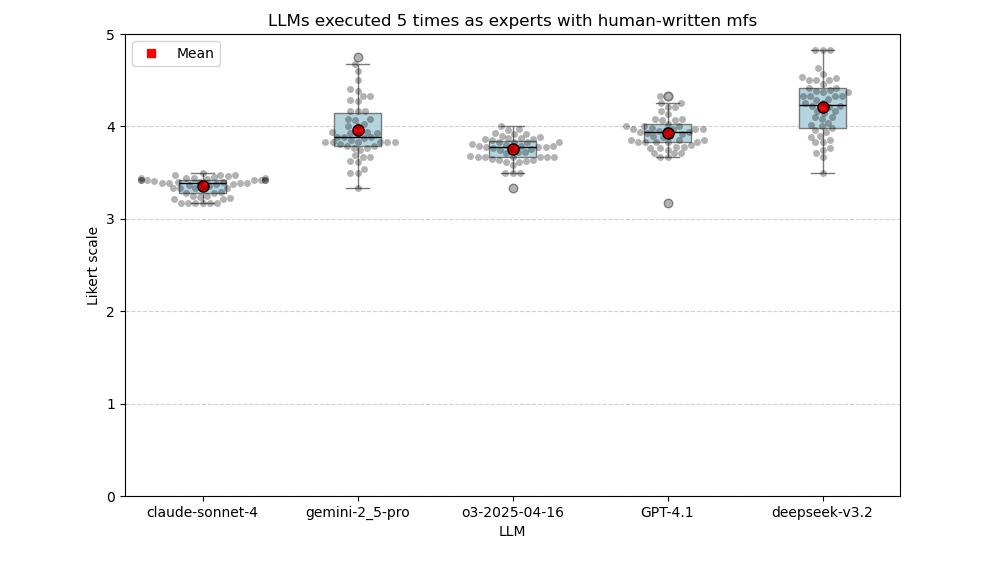}
           \caption{LLMs 
 as literary experts with six human-written~MFs average of five executions. Gray dots represent the AVG values obtained in each MF/LLM execution.}
   \label{fig:llms_excs_hw_mfs_exps}
\end{figure}

    Figure~\ref{fig:llms_excs_hw_mfs_exps} presents the results obtained by testing each LLM with an ``expert'' prompt, using the same human-written microfictions and the GrAImes evaluation protocol. The~results show that the responses from the LLMs across five executions are consistent with the mean value (displayed in red), indicating stability within the~models.

\vspace{-6pt}

          \subsection{Enthusiasts' GrAImes Evaluation Results (Experiment II)}
          
\begin{figure}[H]

       \includegraphics[width=1\textwidth]{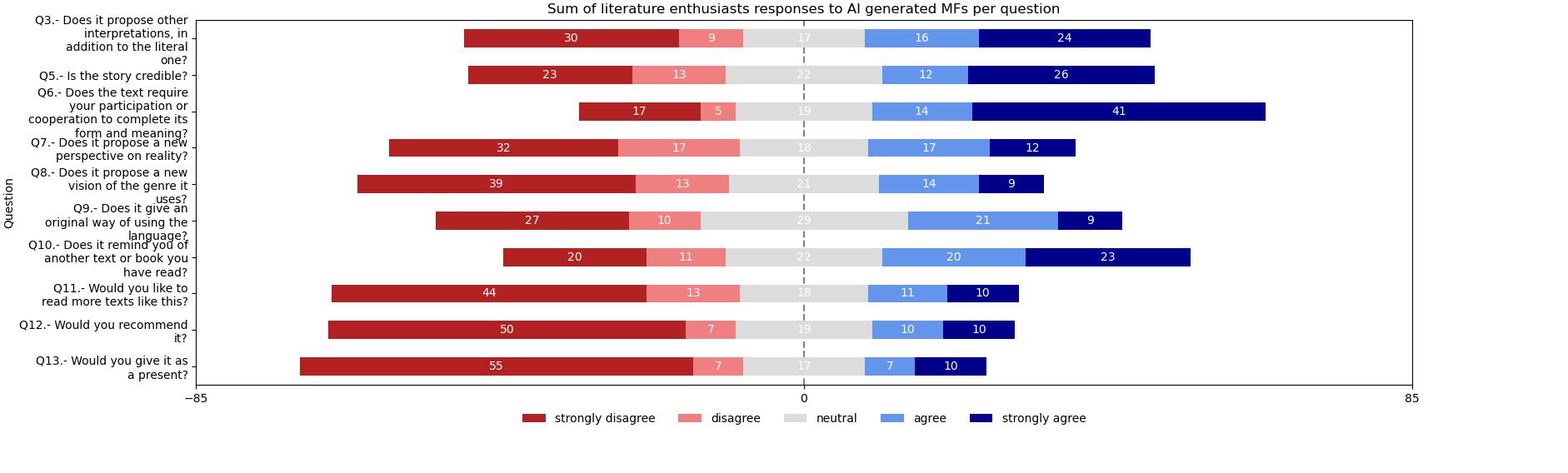}
	    	   \caption{Literature 
 enthusiasts' responses to GrAimes questions with AI-generated~MFs.}
       \label{fig:stacked_bars_likert_enths}
	    
	\end{figure}

\begin{figure}[H]

       \includegraphics[width=1\textwidth]{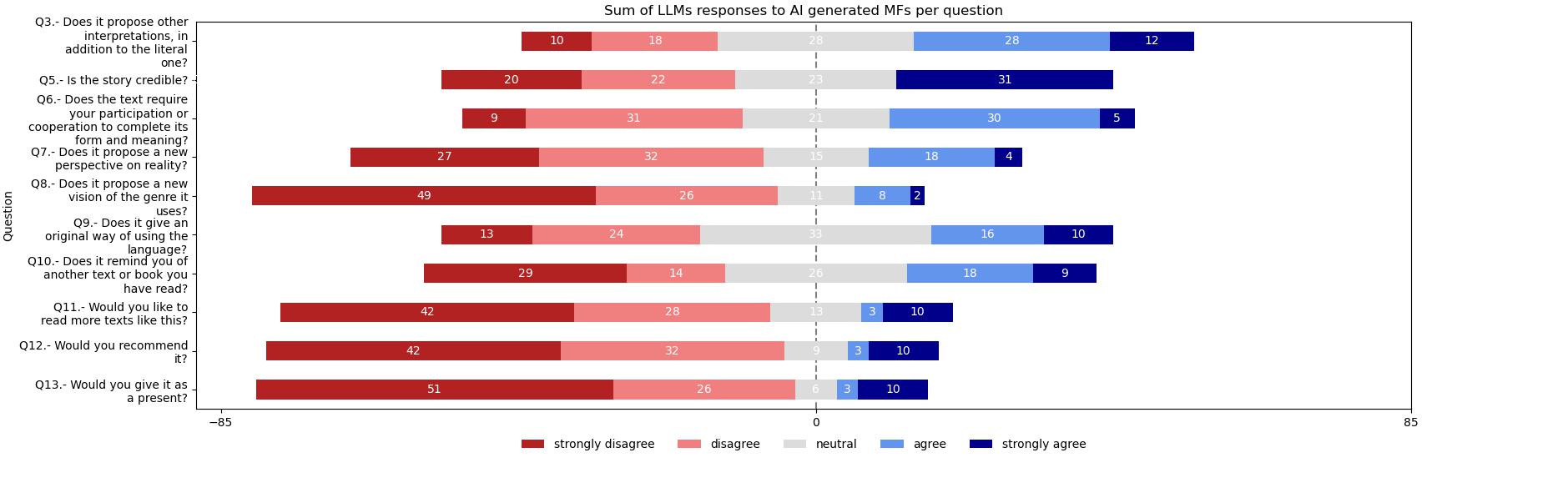}

               \caption{LLMs' 
 responses to GrAimes with AI-generated~microfictions.}
       \label{fig:stacked_bars_likert_llms_enths}
	    
	\end{figure}

\begin{figure}[H]

\vspace{-3pt}
\includegraphics[width=.42\linewidth]{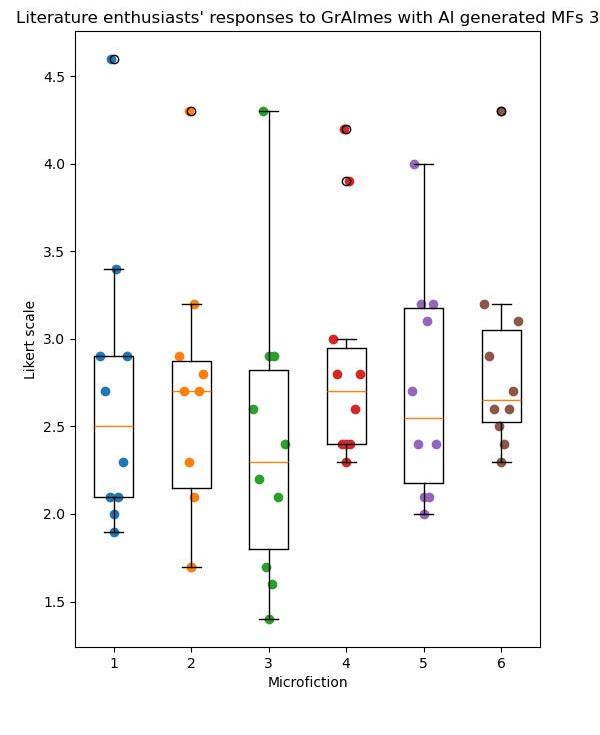}
\includegraphics[width=.5\linewidth]{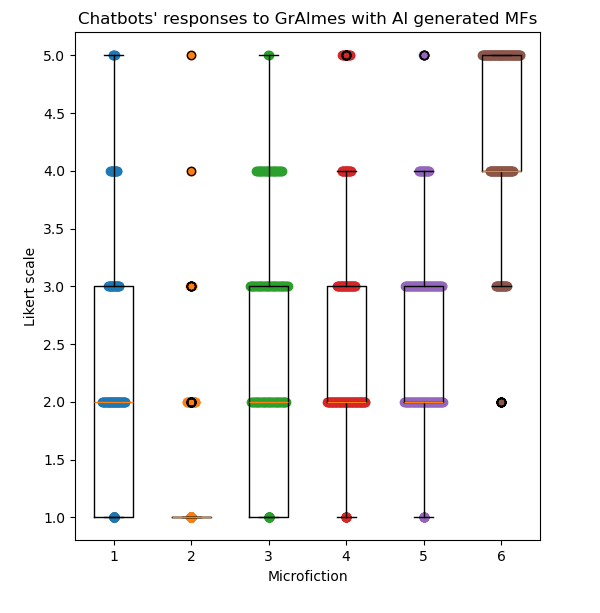}

				\caption{Responses 
 of LLMs using GrAImes to AI-generated~microfictions. Each color represents a different microfiction evaluated.}
				\label{fig:beeswarm_exp_enth_mf_AI}

\end{figure}

        Table~\ref{table:chatbots_ai_mfs} shows the results of the LLM-as-judge assessment of the literary quality of microfictions from the enthusiast dataset (six microfictions generated with Chat-GPT3.5 and Monterroso, as described in Section~\ref{sec:datasets}). The~charts presented in  ~\ref{fig:stacked_bars_likert_enths} ,  ~\ref{fig:stacked_bars_likert_llms_enths} and Figures~\ref{fig:beeswarm_exp_enth_mf_AI} illustrate a comparative analysis of responses from humans and LLMs evaluated using a Likert scale. The~data reveal a significant disparity in perception, with~the literature enthusiasts providing markedly lower average ratings (approximately 2.5) compared to the LLM evaluators' responses, which cluster near the maximum score of~5.0.

This stark contrast indicates that AI-generated microfiction successfully meets the formal and semantic patterns that other AI models are trained to recognize and generate, effectively excelling in a closed-loop AI-to-AI evaluation. Conversely, the~lukewarm reception from human experts highlights that mere pattern replication is insufficient for crafting narratives that resonate on a human level, which involves subjective appreciation along with complex and often implicit aesthetic~criteria. 

\begin{figure}[H]

      \hspace*{-6pt}  \includegraphics[width=.5\linewidth]{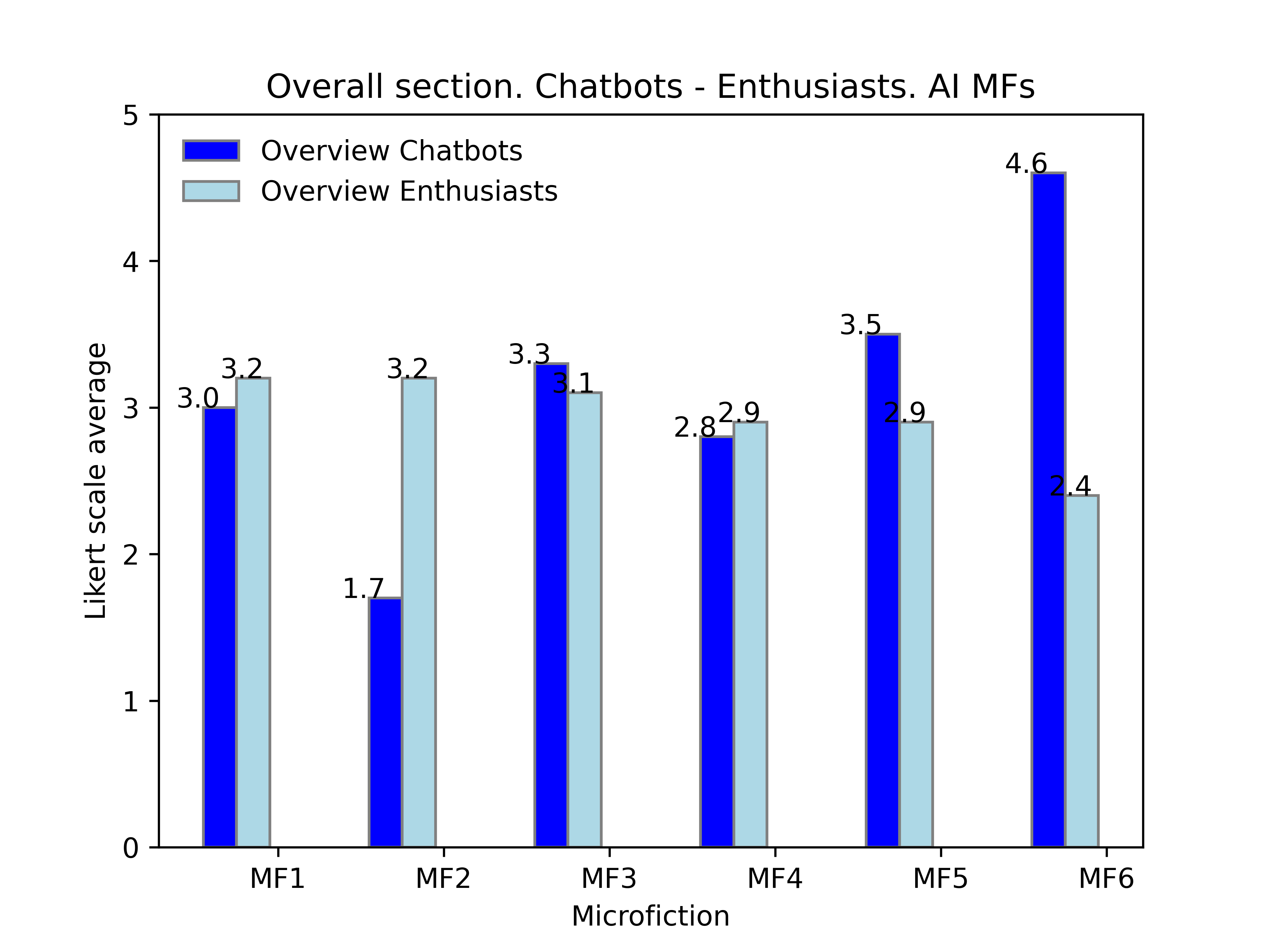}
		\includegraphics[width=.5\linewidth]{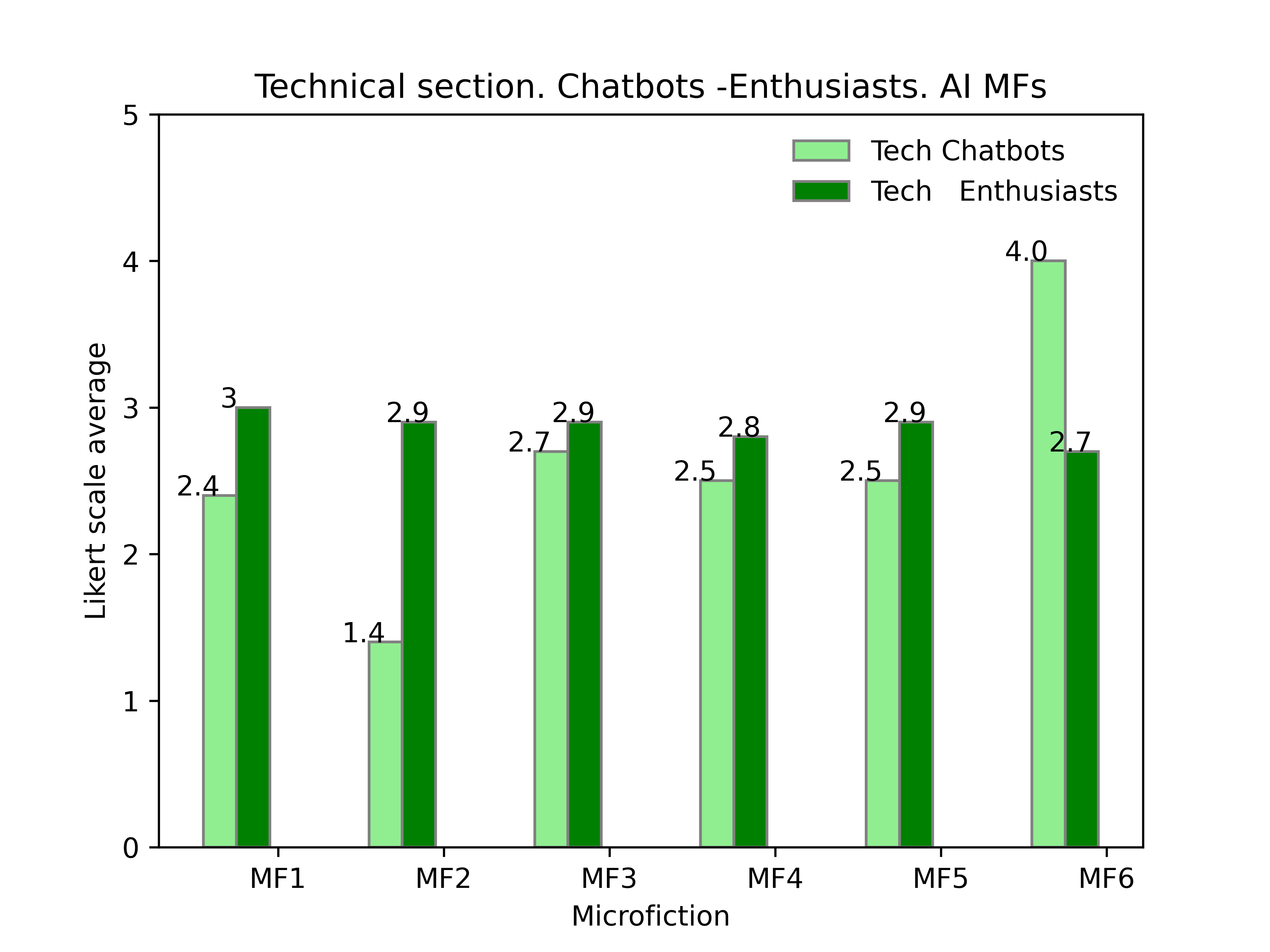}\\
		   \hspace*{-6pt} \includegraphics[width=.5\linewidth]{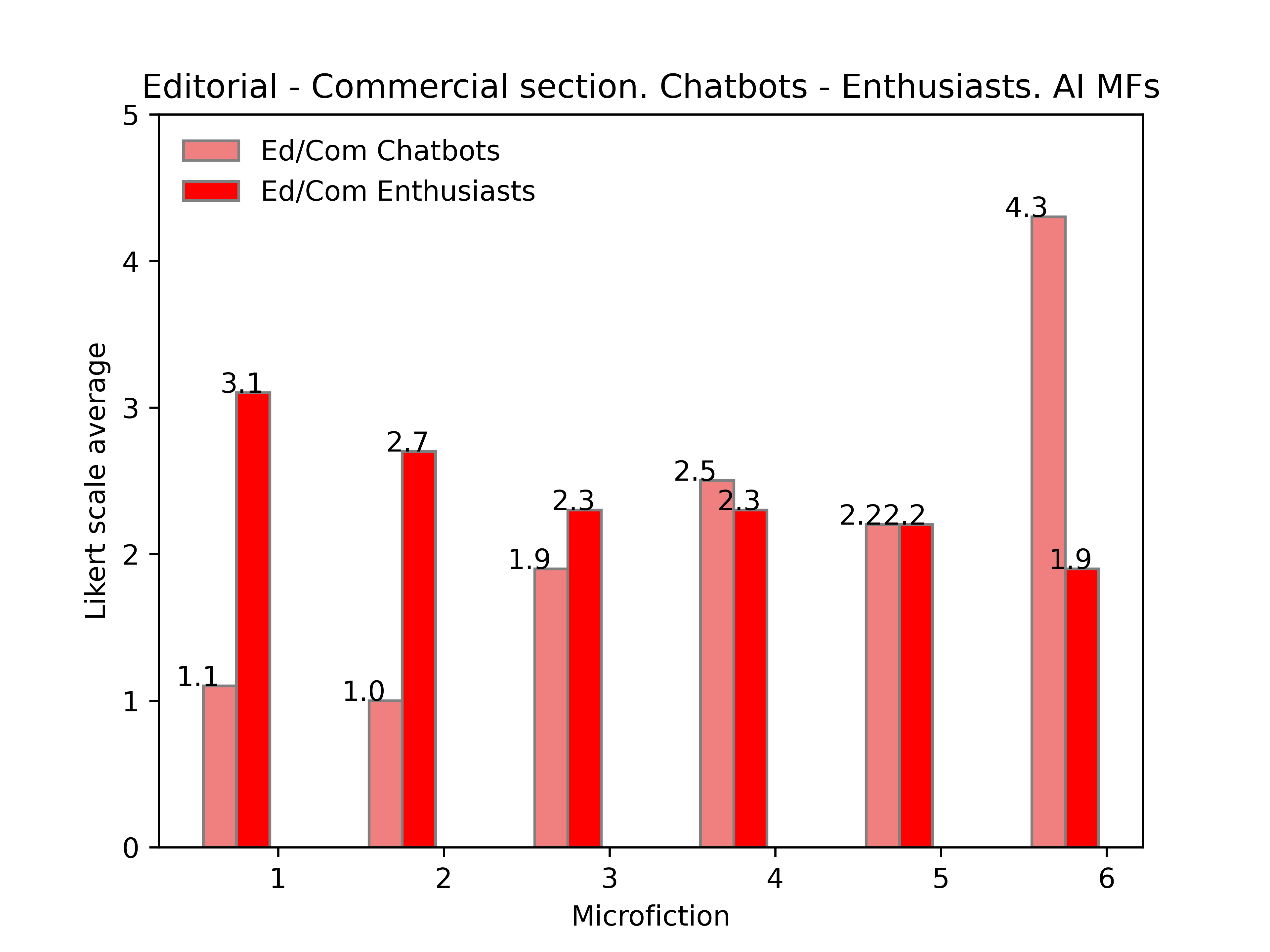}
		\includegraphics[width=.5\linewidth]{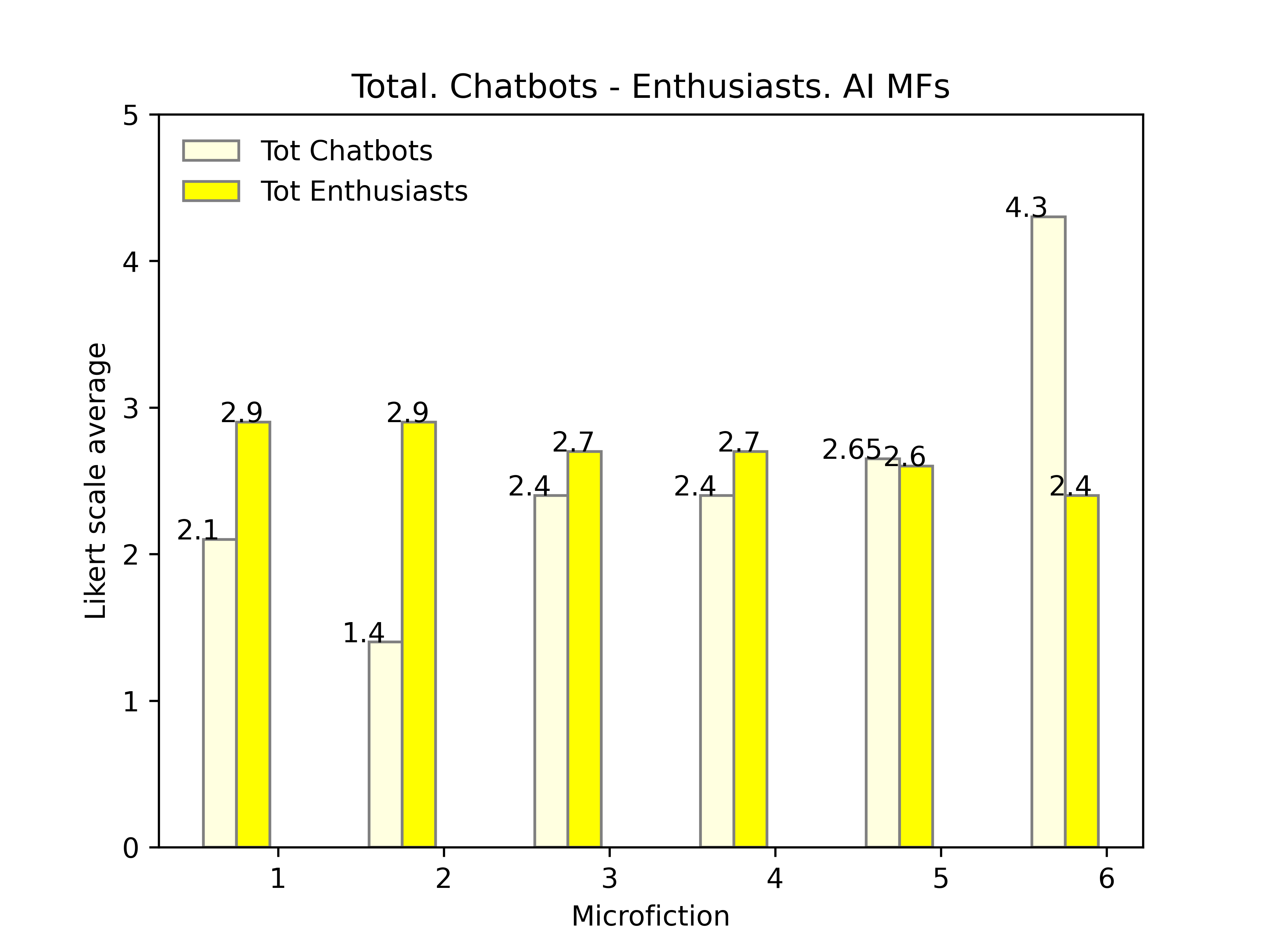}
			\caption{Comparison of ratings of AI-generated~microfiction by LLMs and literature enthusiasts (by GrAImes questionnaire~section).}
\label{fig:comp_exp_enth_mf_AI}
            \end{figure}

\begin{figure}[H]
    \hspace*{-6pt}\includegraphics[width=0.5\linewidth]{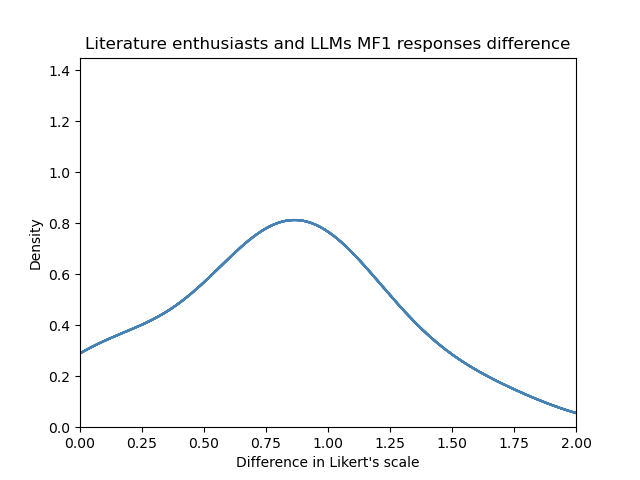}
    \includegraphics[width=0.5\linewidth]{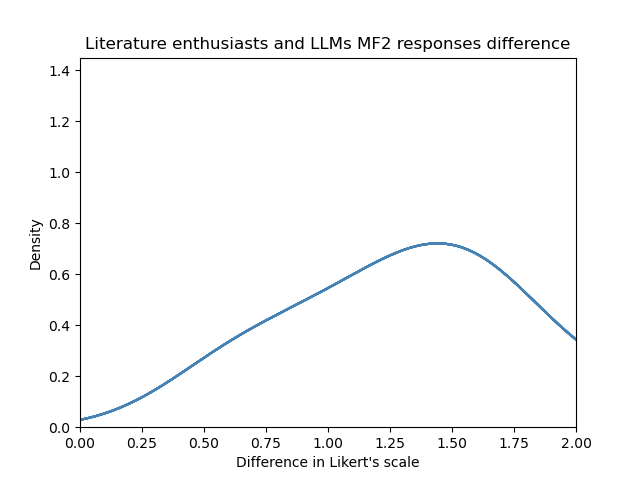}\\
   \hspace*{-6pt} \includegraphics[width=0.5\linewidth]{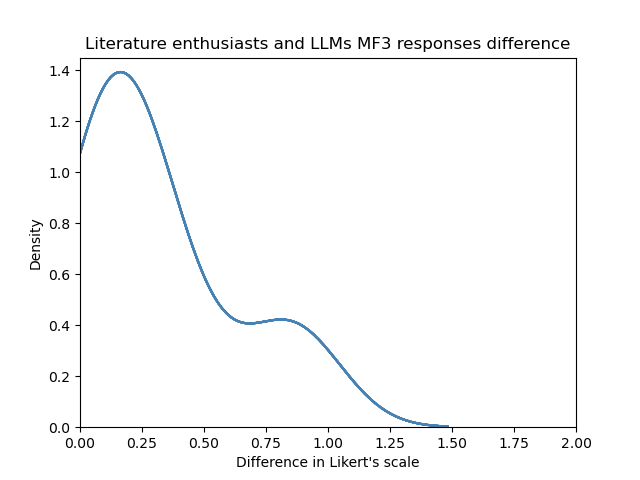}
    \includegraphics[width=0.5\linewidth]{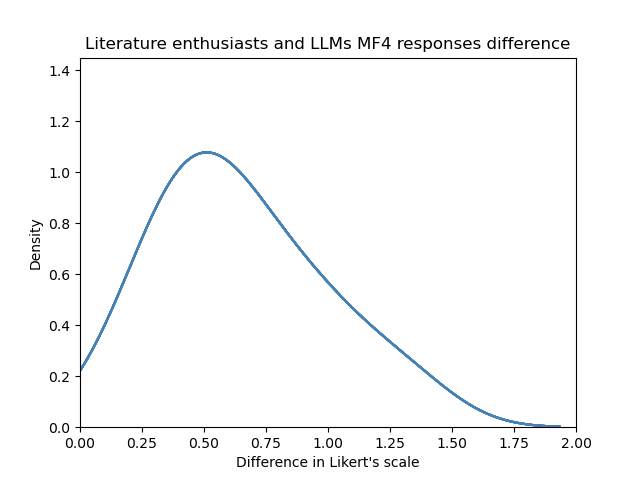}\\
   \hspace*{-6pt} \includegraphics[width=0.5\linewidth]{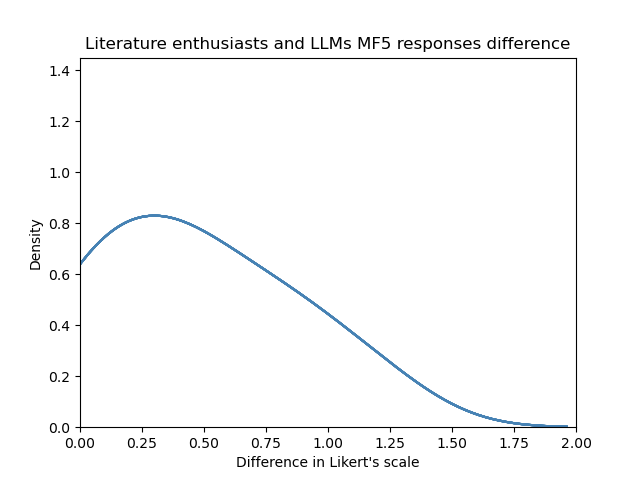}
    \includegraphics[width=0.5\linewidth]{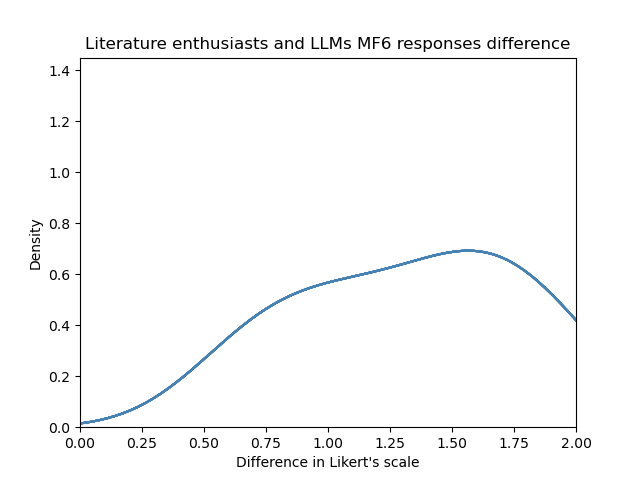}
	    \caption{Differences between literature enthusiasts and LLMs in AV responses (by~microfiction).}
    \label{fig:density_diff_enth}	
\end{figure}

\begin{figure}[H]
\vspace{-12pt}
   \hspace*{-6pt} \includegraphics[width=0.5\linewidth]{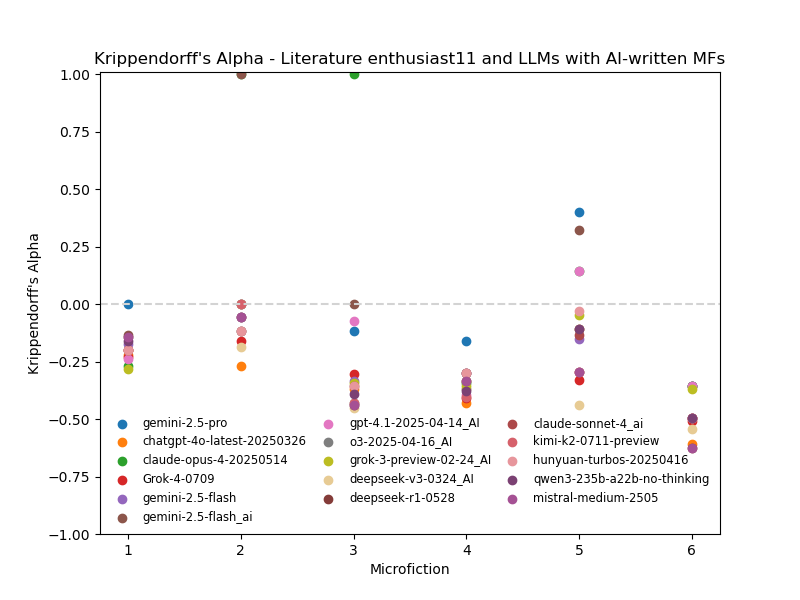}
    \includegraphics[width=0.5\linewidth]{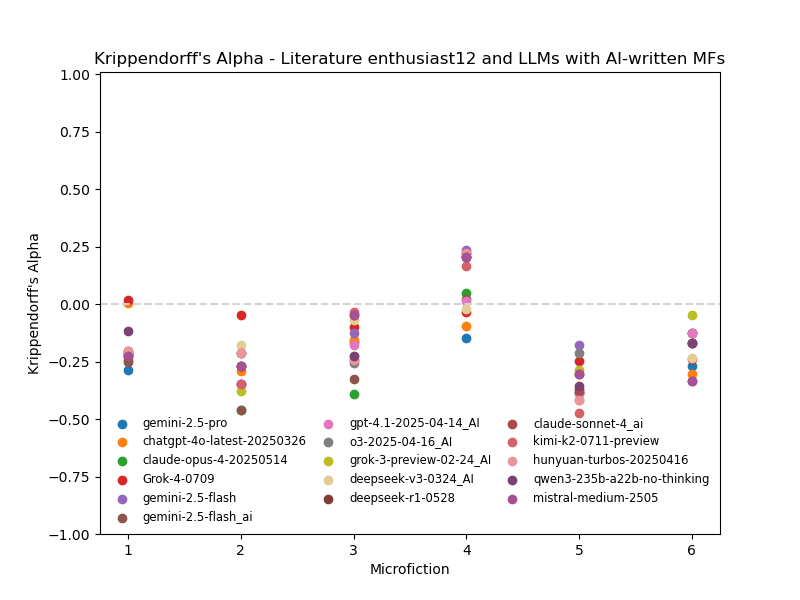}\\
      \hspace*{-6pt} \includegraphics[width=0.5\linewidth]{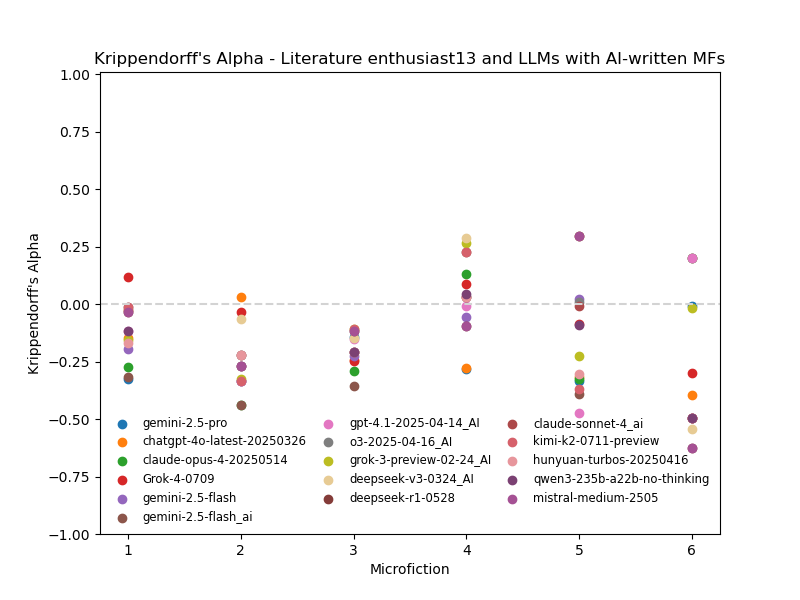}
    \includegraphics[width=0.5\linewidth]{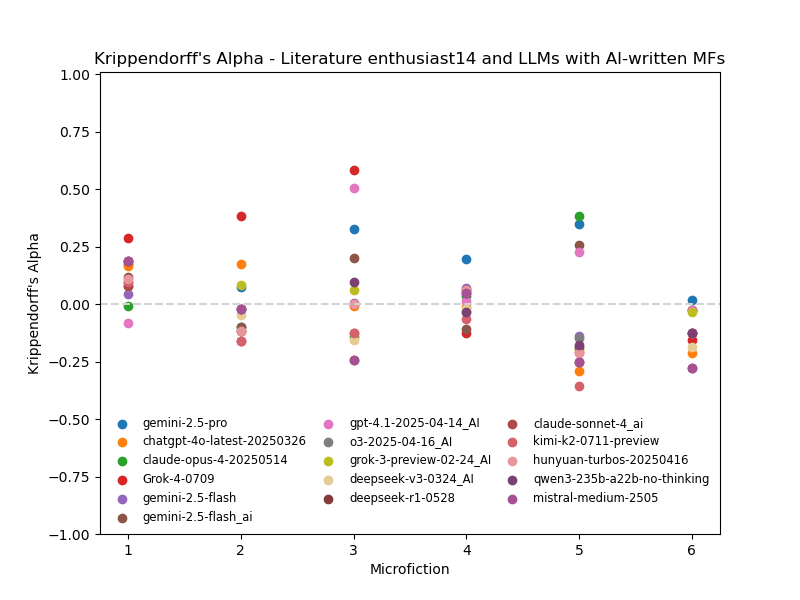}
    \caption{Krippendorff's $\alpha$ for 4 literary enthusiasts and 16 LLMs
.}
    \label{fig:kripendorff_lit_enth16_llms16_1}
\end{figure}

Analyzing the results displayed in Tables~\ref{table:chatbots_ai_mfs} and \ref{table:Chatbots_mf_ai_sects_AV_SD}, we observe that the literature enthusiasts tended to evaluate the microfictions generated by GPT-3.5 more favorably, as~reported in~\cite{aleman2025can}. Similarly, when prompted to act as literature enthusiasts, the LLMs also rated microfictions generated by GPT-3.5 (4--6) more highly than those generated by Monterroso (1--3) as seen in Figures~\ref{fig:comp_exp_enth_mf_AI} and ~\ref{fig:density_diff_enth}. However, the~Krippendorff's $\alpha$ inter-rater reliability analysis for LLMs using the GrAImes protocol with artificial intelligence-generated microfictions from Monterroso and GPT-3.5 indicates that the agreement among the LLMs is no better than chance on the Likert scale responses (see Table~\ref{table:Krippendorf_alpha_le1_llms_ai_mfs} and Figure ~\ref{fig:kripendorff_lit_enth16_llms16_1}).

  
\vspace{-9pt}

\vspace{-9pt}

\begin{table}[H]
	\caption{LLM responses to AI-generated~microfictions (Likert scale).}
	\label{table:chatbots_ai_mfs}
\footnotesize
\begin{adjustwidth}{-\extralength}{0cm}
	\begin{tabular}{
			m{27.8em} m{1.1em}m{1.1em}
			m{1.1em} m{1.1em} m{1.1em} m{1.1em} 
			m{1.1em} m{1.1em} m{1.1em} m{1.1em}
			m{1.1em} m{1.1em} m{1.1em} m{1.1em}
		} 

		\toprule
		\multicolumn{15}{c}{\textbf{LLMs Responses to Human-Written Microfictions}}\\
		\midrule
		& \multicolumn{2}{c}{\textbf{MF1}} 
        & \multicolumn{2}{c}{\textbf{MF2}}
        & \multicolumn{2}{c}{\textbf{MF3}} 
        & \multicolumn{2}{c}{\textbf{MF4}}
        & \multicolumn{2}{c}{\textbf{MF5}} 
        & \multicolumn{2}{c}{\textbf{MF6}} 
        & \multicolumn{2}{c}{\textbf{Avg.}}\\
		\midrule
		\textbf{Question} & \textbf{AV} & \textbf{SD} & \textbf{AV} & \textbf{SD} & \textbf{AV} & \textbf{SD}  & \textbf{AV} & \textbf{SD} & \textbf{AV} & \textbf{SD}& \textbf{AV} & \textbf{SD} & \textbf{AV} & \textbf{SD} \\
		\midrule
		\multicolumn{15}{c}{ Story Overview and Text Complexity} 
\\
		\midrule
		3.-Does it propose other interpretations, in~addition to the literal one? 
		&3.0 & 0.8 & 1.7 & 0.6 & 3.3 & 1.3 & 2.8 & 0.7 & 3.5 & 0.9 & 4.6 & 0.5 & 3.1 & 0.8 \\
				
		\midrule
		\multicolumn{15}{c}{Technical Assessment} 
        \\
		\midrule
		5.-Is the story credible? 
		&2.0 & 0.4 & 1.1 & 0.3 & 2.3 & 0.9 & 4.9 & 0.5 & 2.8 & 0.4 & 5.0 & 0.0 & 3.0 & 0.4 \\

		\midrule
		6.-Does the text require your participation or cooperation to complete its form and meaning?
		&3.8 & 1.0 & 2.1 & 1.1 & 3.4 & 1.2 & 2.0 & 0.4 & 2.7 & 0.6 & 3.6 & 0.6 & 2.5 & 0.8 \\
  		 
		\midrule
		7.-Does it propose a new vision of reality? 
		&2.1 & 1.0 & 1.2 & 0.5 & 3.2 & 1.1 & 1.7 & 0.7 & 2.3 & 0.7 & 3.8 & 0.8 & 2.4 & 0.8 \\

		\midrule
		8.-Does it propose a new vision of the genre it uses?  
		&1.6 & 0.6 & 1.1 & 0.5 & 2.1 & 0.8 & 1.1 & 0.3 & 1.6 & 0.8 & 3.4 & 1.0 & 2.4 & 0.7 \\
	
		\midrule
		9.-Does it propose a new vision of the language itself? 
		&2.6 & 0.8 & 1.6 & 0.7 & 2.3 & 0.9 & 3.0 & 0.7 & 3.3 & 0.9 & 4.3 & 0.9 & 2.9 & 0.8 \\
		\midrule
		\multicolumn{15}{c}{Editorial/Commercial Quality}
        \\	
		\midrule
		10.-Does it remind you of another text or book you have read?
		&1.3 & 0.6 & 1.2 & 0.4 & 2.7 & 0.9 & 3.7 & 0.5 & 2.6 & 0.7 & 4.3 & 0.9 & 2.6 & 0.7 \\

		\midrule
		11.-Would you like to read more texts like this? 
		&1.0 & 0.0 & 1.0 & 0.0 & 1.9 & 0.8 & 2.0 & 0.4 & 2.1 & 0.8 & 4.4 & 0.8 & 2.1 & 0.5 \\
 	
		\midrule 
		12.-Would you recommend it? 
		&1.0 & 0.0 & 1.0 & 0.0 & 1.7 & 0.5 & 2.2 & 0.5 & 2.1 & 0.8 & 4.3 & 1.2 & 2.3 & 0.5 \\

		\midrule
		13.-Would you give it as a present? 
		&1.0 & 0.0 & 1.0 & 0.0 & 1.3 & 0.5 & 1.9 & 0.6 & 1.9 & 0.8 & 4.3 & 1.2 & 2.4 & 0.5 \\

\bottomrule
\end{tabular} 
\end{adjustwidth}
\end{table}

\vspace{-12pt}

\vspace{-15pt}
\unskip

\begin{table}[H]
	\small
	\caption{LLM
 responses using GrAImes sections: summarized AV and SD for AI-generated microfictions.}
	\label{table:Chatbots_mf_ai_sects_AV_SD}
\begin{adjustwidth}{-\extralength}{0cm}

	\begin{tabularx}{\fulllength}{CCCCCCCCCCCCCCCC} 
		\toprule
		\multicolumn{4}{c}{\textbf{Story Overview}} & & \multicolumn{3}{c}{\textbf{Technical Assessment}} & &  \multicolumn{3}{c}{\textbf{Editorial/Commercial}} & & \multicolumn{3}{c}{\textbf{Total Analysis}}  \\
		\cmidrule{1-4}\cmidrule{6-8}\cmidrule{10-12}\cmidrule{14-16}
		\#& \textbf{MF} & \textbf{AV} & \textbf{SD} && \textbf{MF} & \textbf{AV} & \textbf{SD} & & \textbf{MF} & \textbf{AV} & \textbf{SD} & & \textbf{MF} & \textbf{AV} & \textbf{SD} \\ 
		\midrule
		1 & 6 & 4.6 & 0.5 && 6 & 4.0 & 0.7 &&  6 & 4.3 & 1.0 && 6 & 4.2 & 0.8 \\
				\midrule
		2 & 5 & 3.5 & 0.9 && 3 & 2.7 & 1.0 &&  4 & 2.5 & 0.5 && 4 & 2.5 & 0.5 \\
				\midrule
		3 & 3 & 3.3 & 1.3 && 4 & 2.5 & 0.5 &&  5 & 2.2 & 0.8 && 5 & 2.5 & 0.7 \\
				\midrule
		4 & 1 & 3.0 & 0.8 && 5 & 2.5 & 0.7 && 3 & 1.9 & 0.7 && 3 & 2.4 & 0.9 \\
				\midrule
		5 &4 & 2.8 & 0.7 && 1 & 2.4 & 0.7 && 1 & 1.1 & 0.1 && 1 & 1.9 & 0.5 \\
				\midrule
		6 & 2 & 1.7 & 0.6 && 2 & 1.4 & 0.6 && 2 & 1.0 & 0.1 && 2 & 1.3 & 0.4\\		
		\bottomrule
%
	\end{tabularx}
\end{adjustwidth}
\end{table}
\vspace{-6pt}


\vspace{-6pt}

\begin{table}[H]
\caption{Krippendorff's $\alpha$ and CI of literature enthusiasts and LLMs for AI-generated microfictions. 
}
\label{table:Krippendorf_alpha_le1_llms_ai_mfs}
\scriptsize
\begin{adjustwidth}{-\extralength}{0cm}

	\begin{tabular}{p{.2cm} p{.54cm} p{1.59cm} 
    p{.54cm} p{1.59cm} 
    p{.54cm} p{1.59cm} 
    p{.54cm} p{1.59cm} 
    p{.54cm} p{1.59cm} 
    p{.54cm} p{1.59cm} }
		\toprule
		\multicolumn{13}{c}{\textbf{Krippendorff's \boldmath{$\alpha$} and Confidence Interval (CI) for Literature Enthusiast 1 (16 LLMs)}}  \\	
		\midrule
            \textbf{LLM} & \textbf{MF1}&& \textbf{MF2}&& \textbf{MF3}&&\textbf{MF4}&& \textbf{MF5}&&\textbf{MF6}& \\ 
		\midrule
		 & \boldmath{$\alpha$}&\textbf{CI}& \boldmath{$\alpha$} & \textbf{CI}& \boldmath{$\alpha$} & \textbf{CI}& \boldmath{$\alpha$} & \textbf{CI}& \boldmath{$\alpha$}  & \textbf{CI}& \boldmath{$\alpha$} & \textbf{CI} \\ 
		\midrule

1&$-$0.348 & (0.268, 0.605)& $-$0.188 & (0.106, 0.500)&$-$0.078 & (0.054, 0.376)&$-$0.013& (0.062, 0.443)&$-$0.163& (0.135, 0.564)&$-$0.126& (0.095, 0.476)\\ 
        \midrule
2 &0.007& ($-$0.019, 0.210)&0.125& ($-$0.022, 0.185)&0.250& ($-$0.032, 0.213)&0.095& (0.014, 0.397)&$-$0.163& (0.161, 0.476)&0.021& ($-$0.040, 0.419)\\ 
        \midrule
3 &$-$0.242& (0.139, 0.399)&$-$0.299& (0.192, 0.551)&$-$0.299& (0.195, 0.578)&$-$0.125& (0.087, 0.408)&$-$0.242& (0.234, 0.488)&$-$0.163& (0.087, 0.430)\\ 
        \midrule
4 &$-$0.082& (0.007, 0.282)&0.255& ($-$0.018, 0.243)&0.265& ($-$0.019, 0.311)&0.070& ($-$0.002, 0.369)&0.143& ($-$0.013, 0.291)&0.128& ($-$0.030, 0.361)\\ 
        \midrule
5 &$-$0.242& (0.080, 0.355)&$-$0.188& (0.115, 0.501)&0.395& ($-$0.045, 0.185)&$-$0.007& (0.026, 0.378)&$-$0.027& (0.054, 0.434)&0.208& ($-$0.056, 0.324)\\ 
        \midrule
6  &$-$0.242& (0.101, 0.484)&$-$0.299& (0.192, 0.556)&$-$0.197& (0.100, 0.529)&$-$0.226& (0.147, 0.441)&$-$0.218& (0.225, 0.476)&$-$0.163& (0.087, 0.434)\\ 
        \midrule
7 &$-$0.195& (0.089, 0.365)&$-$0.188& (0.128, 0.488)&$-$0.063& (0.034, 0.390)&$-$0.148& (0.103, 0.531)&$-$0.242& (0.234, 0.488)&$-$0.163& (0.087, 0.476)\\ 
        \midrule
8 &$-$0.195& (0.059, 0.316)&$-$0.056& (0.076, 0.345)&0.119& ($-$0.020, 0.235)&0.007& (0.029, 0.378)&$-$0.163& (0.076, 0.373)&0.208& ($-$0.056, 0.340)\\ 
		\midrule
9 &$-$0.242& (0.087, 0.397)&$-$0.216& (0.084, 0.500)&0.019& ($-$0.001, 0.276)&$-$0.007& (0.038, 0.378)&$-$0.242& (0.234, 0.488)&0.228& ($-$0.046, 0.232)\\ 
        \midrule
10&$-$0.195& (0.053, 0.349)&$-$0.020& (0.067, 0.335)&$-$0.125& ($-$0.001, 0.336)&0.113& (0.001, 0.365)&$-$0.221& (0.104, 0.531)&$-$0.044& ($-$0.056, 0.467)\\ 
        \midrule
11&$-$0.203& (0.071, 0.393)&$-$0.109& (0.095, 0.419)&0.095& ($-$0.019, 0.183)&$-$0.007& (0.045, 0.489)&$-$0.221& (0.105, 0.486)&$-$0.357& ($-$0.056, 0.648)\\ 
        \midrule
12&$-$0.075& (0.043, 0.297)&$-$0.056& (0.073, 0.360)&0.255& ($-$0.033, 0.183)&0.007& (0.034, 0.365)&$-$0.267& (0.144, 0.430)&0.208& ($-$0.056, 0.340)\\ 
        \midrule
13&$-$0.195& (0.057, 0.325)&$-$0.188& (0.126, 0.500)&$-$0.132& ($-$0.006, 0.248)&$-$0.140& (0.076, 0.488)&$-$0.284& (0.296, 0.578)&0.208& ($-$0.056, 0.340)\\ 
        \midrule
14&$-$0.218& (0.057, 0.287)&$-$0.056& (0.077, 0.356)&0.255& ($-$0.035, 0.159)&0.007& (0.034, 0.397)&$-$0.242& (0.242, 0.476)&0.208& ($-$0.056, 0.324)\\ 
        \midrule
15&$-$0.089& (0.054, 0.388)&$-$0.109& (0.095, 0.438)&0.007& ($-$0.008, 0.258)&0.101& (0.015, 0.442)&$-$0.179& (0.077, 0.441)&0.208& ($-$0.056, 0.340)\\ 
        \midrule
                16& $-$0.203 & (0.071, 0.393)& $-$0.109  & (0.086, 0.419)& 0.095 & ($-$0.016, 0.205)& $-$0.007 & (0.037, 0.461)& $-$0.221 & (0.104, 0.501)& 0.357&  ($-$0.056, 0.648) \\ 
\bottomrule
\end{tabular} 
\end{adjustwidth}
\end{table}

\subsubsection*{LLM Benchmark on the GrAImes Enthusiasts Dataset (Experiment II) 
}

The data presented in Table~\ref{table:LLMs_avg_closer_to_enth} rank the sixteen LLMs based on the average absolute divergence of their evaluative scores from those of human literature enthusiasts across ten distinct questions (Q3--Q13) evaluating microfictions. The~key metric is the final ``Avg” column, which represents the mean of the absolute differences for each model. 
The top-performing models, including deepseek-r1-0528 and mistral-medium-2505, achieve a near-zero average divergence (Avg = 0.1), suggesting that these models have been optimized or architected in a way that successfully captures the statistical and evaluative patterns deemed significant by a human~benchmark.

Conversely, the~lowest-ranked models, such as gemini-2.5-pro and claude-opus-4-20250514 (Avg = 0.9--1.0), exhibit a consistent negative divergence across the vast majority of evaluative criteria. This systematic and directional deviation suggests a fundamental misalignment between their internal evaluative frameworks and the human benchmark. Their scoring behavior indicates a potential bias towards different latent factors, possibly prioritizing logical coherence, syntactic complexity, or~other non-stylistic features over the narrative and aesthetic preferences demonstrated by human~raters.

\begin{table}[H]
	\small
		\caption{LLMs ordered by the average values that are closer to the enthusiasts average responses (by GrAImes Likert scale~question).}
	\label{table:LLMs_avg_closer_to_enth}

\begin{adjustwidth}{-\extralength}{0cm}
\begin{tabular}{c m{5.5 cm} c c c c c c c c c c c }
	\toprule
	\textbf{\# 
} & \textbf{LLM} & \textbf{Q3} & \textbf{Q5} & \textbf{Q6} & \textbf{Q7} & \textbf{Q8} & \textbf{Q9} & \textbf{Q10} & \textbf{Q11} & \textbf{Q12} & \textbf{Q13} & \textbf{Avg} \\
	\midrule
	1 & deepseek-r1-0528 & 0.2 & 0.1 & $-$0.4 & $-$0.1 & $-$0.2 & 0.6 & $-$0.2 & 0.2 & 0.1 & 0.3 & 0.1 \\
	\midrule
	2 & mistral-medium-2505 & 0.2 & 0.1 & $-$0.4 & $-$0.1 & $-$0.2 & 0.6 & $-$0.2 & 0.2 & 0.1 & 0.3 & 0.1 \\
	\midrule
	3 & o3-2025-04-16\_AI & 0.6 & 0.1 & $-$0.4 & $-$0.3 & $-$0.6 & 0.6 & $-$0.3 & $-$0.1 & $-$0.1 & 0.1 & 0.1 \\
	\midrule
	4 & deepseek-v3-0324\_AI & 0.6 & $-$0.1 & $-$0.6 & 0.4 & $-$0.2 & 0.6 & $-$0.2 & 0.2 & 0.3 & 0.5 & 0.1 \\
	\midrule
	5 & kimi-k2-0711-preview & 0.2 & $-$0.1 & $-$0.6 & 0.1 & $-$0.2 & 0.1 & $-$0.7 & 0.1 & 0.1 & 0.1 & 0.1 \\
	\midrule
	6 & claude-sonnet-4\_ai & 0.6 & 0.3 & $-$0.4 & $-$0.3 & $-$0.6 & 0.4 & $-$0.5 & $-$0.1 & $-$0.1 & 0.0 & 0.1 \\
	\midrule
	7 & gemini-2.5-flash & 1.2 & 0.1 & $-$0.6 & $-$0.3 & $-$0.7 & 0.3 & $-$0.3 & 0.1 & 0.1 & 0.0 & 0.2 \\
	\midrule
	8 & hunyuan-turbos-20250416 & 0.6 & 0.1 & $-$0.4 & $-$0.3 & $-$0.6 & 0.4 & $-$0.5 & $-$0.1 & $-$0.1 & 0.0 & 0.2 \\
	\midrule
	9 & chatgpt-4o-latest-20250326 & 0.6 & 0.3 & 0.2 & 1.1 & 0.3 & 0.9 & $-$0.5 & $-$0.3 & $-$0.2 & $-$0.2 & 0.2 \\
	\midrule
	10 & qwen3-235b-a22b-no-thinking & 0.6 & $-$0.1 & $-$0.3 & $-$0.3 & $-$0.7 & 0.3 & $-$0.5 & $-$0.1 & $-$0.1 & 0.0 & 0.2 \\
	\midrule
	11 & grok-4-0709 & 0.6 & $-$0.2 & 0.1 & 0.6 & 0.1 & 1.1 & 0.2 & $-$0.1 & $-$0.1 & 0.1 & 0.2 \\
	\midrule
	12 & grok-3-preview-02-24\_AI & $-$0.6 & $-$0.1 & $-$0.4 & $-$0.4 & $-$0.7 & $-$0.2 & $-$0.3 & $-$0.1 & 0.1 & $-$0.2 & 0.3 \\
	\midrule
	13 &  gemini-2.5-pro & $-$1.1 & $-$0.4 & $-$1.8 & $-$1.3 & $-$1.2 & $-$1.1 & $-$0.7 & $-$0.8 & $-$0.7 & $-$0.4 & 0.9 \\
	\midrule
	14 & gpt-4.1-2025-04-14\_AI & $-$0.1 & $-$0.4 & $-$1.4 & $-$0.6 & $-$1.1 & $-$0.9 & $-$1.3 & $-$0.8 & $-$0.9 & $-$0.9 & 0.9 \\
	\midrule
	15 &  claude-opus-4-20250514 & $-$0.8 & $-$0.2 & $-$1.8 & $-$0.9 & $-$1.1 & $-$0.9 & $-$1.3 & $-$0.8 & $-$0.9 & $-$0.9 & 1.0 \\
	\midrule
	16 & gemini-2.5-flash\_ai & 0.1 & $-$0.6 & $-$1.8 & $-$0.9 & $-$1.2 & $-$0.9 & $-$1.3 & $-$0.8 & $-$0.9 & $-$0.9 & 1.0 \\
	\bottomrule
\end{tabular}

\end{adjustwidth}	
\end{table}

\subsection{TTCW and GrAImes Protocol Comparison Results (Experiment III)}

    To test the GrAImes evaluation protocol, we used the TTCW~\citep{chakrabarty2024art} experiments dataset. This dataset is composed of 12 short stories written by humans and 36 generated by three different LLMs. To~comply with the yes/no answer scale used in TTCW~\citep{chakrabarty2024art}, we considered our Likert scale (1--5) as follows: 1--2 = No, 4--5 = Yes.~We proportionally distributed scores of 3 according to the average yes/no answers to all the GrAImes questions. We then used an Average Passing Rate (APR) measuring the number of Yes responses obtained on each~question: 
    \[y = \frac{\sum_{i=1}^{n} x = 1 +  \sum_{i=1}^{n} x = 2}{n}\]
        
        \[z = \frac{\sum_{i=1}^{n} x = 4 +  \sum_{i=1}^{n} x = 5}{n}\]

         \[t = \frac{\sum_{i=1}^{n} x = 3}{n} \]

        \[a = y *(1+ \frac{t}{1-t}) ; \text{(distributing to 0 the percentage of \textit{t})}\] 

        \[b = z *(1+ \frac{t}{1-t}) ; \text{(distributing to 1 the percentage of \textit{t})}\] 

        \[\text{If}~a > b~\text{then}~0; \text{else}~1\]
where 
 \textit{n} is the quantity of short stories (12), \textit{x} is the Likert value, \textit{y} is the average of \textit{x} values equal to 1 or 2, \textit{z} is the average \textit{x} values equal to 4 or 5, and~\textit{t} is the average of \textit{x} 
 values equal to 3. Then, the value of \textit{a} is the percentage added to 0 from the percentage distribution of \textit{x} = 3 and~\textit{b} is the percentage added to 1
 from the percentage distribution of \textit{x} = 3. Finally, if~\textit{a} is greater than \textit{b}, then the final value is 0; otherwise, it is~1.


    From Tables~\ref{table:llms_ai_shst} and  
    \ref{table:llms_ai_shst_claude}, it can be seen that LLMs evaluate the short stories written by humans (published in The New Yorker magazine) (AV = 4.27\%) as being better than the fictions generated by AI. We also observe that the most recent LLMs tended to rate short stories higher than those used in the original TTCW experiment~\citep{chakrabarty2024art}. This could be due to modifications in training and programming to facilitate empathic responses \citep{welivita2024large}.

    The data from the three LLMs (GPT-5-high, Claude-opus-4-1, and~ChatGPT-4o-latest) reveals a consistent and statistically significant evaluative hierarchy based on the source of the short stories. Across all models, stories sourced from The New Yorker consistently received the highest average scores (ranging from 4.11 to 4.62), followed by those from GPT-4 (3.68 to 4.25). Stories generated by GPT-3.5-turbo and the earlier Claude V1.3 model were consistently rated the~lowest. In Tables~\ref{table:llms_ai_sht_yn} and \ref{table:llms_ai_sht_yn_claude} we can see the Y/N and APR of LLMs GPT-5-high and Claude-opus-4-1-20250805.

\begin{table}[H]
	\small
	\caption{GPT-5-high LLM AV responses to Likert-scale questions on all 48 short stories by source (The New Yorker, Claude V1.3, GPT-3.5-turbo, and GPT-4).}
    \label{table:llms_ai_shst}
\begin{adjustwidth}{-\extralength}{0cm}

	\begin{tabularx}{\fulllength}{
			m{28em} c c  c  c  c  c
		} 
		\midrule
		\multicolumn{7}{c}{\textbf{GPT-5-high LLM Responses to Short Stories (Likert Scale 1--5)}}\\
		\midrule 
        &\multicolumn{1}{c}{\textbf{The New Yorker}}
	& \multicolumn{1}{c}{\textbf{Claude}} & \multicolumn{1}{c}{\textbf{GPT-3.5-turbo}} & \multicolumn{1}{c}{\textbf{GPT-4}} & \multicolumn{1}{c}{\textbf{ALL}}\\
		\midrule
		\textbf{Question} & \textbf{AV} & \textbf{AV} & \textbf{AV} & \textbf{AV}&   \textbf{AV}& \\
		\midrule
		\multicolumn{7}{c}{Story Overview and Text Complexity 
}\\
		\midrule
		3.-Does it propose other interpretations, in~addition to the literal one? 
		& 5 & 4.08 & 4.25 & 5 & 4.53&\\
		\midrule
		\multicolumn{7}{c}{Technical Assessment}
         \\
		\midrule
		5.-Is the story credible? 
		& 4.73 &  4.42&  4 & 4.17 & 4.26&\\

		\midrule
		6.-Does the text require your participation or cooperation to complete its form and meaning?
		& 4.09 & 2.92 & 3.42 & 4.83 & 3.62&\\
  		 
		\midrule
		7.-Does it propose a new vision of reality? 
		& 4 & 3.08 & 3.25 & 3.92 & 3.36&\\

		\midrule
		8.-Does it propose a new vision of the genre it uses?  
		& 2.91 & 2.25 & 2.58 & 3.42 & 2.6&\\
	
		\midrule
		9.-Does it propose a new vision of the language itself? 
		& 4.18 & 2.42 & 2.75 & 4.17 & 3.17&\\

		\midrule
		\multicolumn{7}{c}{Editorial/Commercial Quality}
         \\	
		\midrule
		10.-Does it remind you of another text or book you have read?
		& 4.18 & 3.75 & 3.83 & 4.08 & 3.96&\\

		\midrule
		11.-Would you like to read more texts like this? 
		& 4.82 & 3.92 & 3.92 & 4.83 & 4.15&\\
 	
		\midrule 
		12.-Would you recommend it? 
		& 4.91 & 3.83 & 3.92 & 4.42 & 4.13&\\

		\midrule
		13.-Would you give it as a present? 
		& 3.91 & 3.50 & 3.25 & 3.67 & 3.45&\\
	
		\midrule

        Total by short story source & 4.27 
         & 3.42
         & 3.52 
        & 4.25
        & 3.72
        &\\
        \bottomrule
	\end{tabularx}
\end{adjustwidth}
\end{table}
\vspace{-6pt}

\begin{table}[H]
		\small	
	\caption{Claude-opus-4-1-20250805 LLM AV responses to Likert-scale questions on all 48 short stories by source (The New Yorker, Claude V1.3, GPT-3.5-turbo and GPT-4).}
    \label{table:llms_ai_shst_claude}
\begin{adjustwidth}{-\extralength}{0cm}

	\begin{tabularx}{\fulllength}{
			m{28em} c c  c  c  c  c
		} 
		\midrule
		\multicolumn{7}{c}{\textbf{Claude-Opus-4-1-20250805 LLM Responses to Short Stories (Likert Scale 1--5)}}\\
		\midrule 
        &\multicolumn{1}{c}{\textbf{The New Yorker}}
	& \multicolumn{1}{c}{\textbf{Claude}} & \multicolumn{1}{c}{\textbf{GPT-3.5-turbo}} & \multicolumn{1}{c}{\textbf{GPT-4}} & \multicolumn{1}{c}{\textbf{ALL}}\\
		\midrule
		\textbf{Question} & \textbf{AV} & \textbf{AV} & \textbf{AV} & \textbf{AV}&   \textbf{AV}& \\
		\midrule
		\multicolumn{7}{c}{ Story Overview and Text Complexity}
         \\
		\midrule
		3.-Does it propose other interpretations, in~addition to the literal one? 
		& 4.91 & 3.5 & 2.83 & 4.25 & 3.87 &\\
		\midrule
		5.-Is the story credible? 
		& 4.45 & 3.58 & 2.42 & 3.75 & 3.55 &\\

		\midrule
		6.-Does the text require your participation or cooperation to complete its form and meaning?
		& 4.55 & 2.67 & 2.08 & 3.83 & 3.28 &\\
  		 
		\midrule
		7.-Does it propose a new vision of reality? 
		& 4 & 2.25 & 1.83 & 3.25 & 2.83&\\

		\midrule
		8.-Does it propose a new vision of the genre it uses?  
		& 3.55 & 1.75 & 1.5 & 3.25 & 2.51 &\\
	
		\midrule
		9.-Does it propose a new vision of the language itself? 
		& 3.82 & 2.08 & 1.75 & 3.92 & 2.89 &\\

		\midrule
		\multicolumn{7}{c}{Editorial/Commercial Quality} 
         \\	
		\midrule
		10.-Does it remind you of another text or book you have read?
		& 4 & 4.08 & 4.25 & 3.58 & 3.98 &\\

		\midrule
		11.-Would you like to read more texts like this? 
		& 4.27 & 2.75 & 2.08 & 3.92 & 3.26 &\\
 	
		\midrule 
		12.-Would you recommend it? 
		& 4.27 & 2.67 & 2.25 & 3.92 & 3.28 &\\

		\midrule
		13.-Would you give it as a present? 
		& 3.27 & 2 & 2 & 3.08 & 2.59 &\\
	
		\midrule

        Total by short story source & 4.11 
        & 2.73 
        & 2.3
        & 3.68
        & 3.2
        &\\

		\bottomrule
\end{tabularx} 
\end{adjustwidth}
\end{table}

\vspace{-6pt}

    \begin{table}[H]
		\small	
	\caption{GPT-5-high 
 LLM AV and APR responses to questions on Yes/No  (Y/N) scale for all 48 short stories by source (The New Yorker, Claude V1.3, GPT-3.5-turbo and GPT-4). }
    \label{table:llms_ai_sht_yn}
\begin{adjustwidth}{-\extralength}{0cm}

	\begin{tabularx}{\fulllength}{
			m{26em} c c  c  c  c  c c c
		} 
		\midrule
		\multicolumn{9}{c}{\textbf{GPT-5-high LLM Responses to Short Stories  1/0 (Y/N)}}\\
		\midrule 
        &\multicolumn{2}{c}{\textbf{The New Yorker}}
	& \multicolumn{2}{c}{\textbf{Claude}} & \multicolumn{2}{c}{\textbf{GPT-3.5-turbo}} & \multicolumn{2}{c}{\textbf{GPT-4}}\\
		\midrule
		\textbf{Question} & \textbf{Y/N} & \textbf{APR}& \textbf{Y/N} &\textbf{APR}& \textbf{Y/N} &\textbf{APR}& \textbf{Y/N}&\textbf{APR}\\
		\midrule
		\multicolumn{9}{c}{ Story Overview and Text Complexity}
        \\
		\midrule
		3.-Does it propose other interpretations, in~addition to the literal one? 
        &  1& 100\%& 1 & 91.7\%& 1 & 100\%& 1 & 100\%\\
	
		\midrule
		\multicolumn{9}{c}{Technical Assessment}
        \\
		\midrule
		5.-Is the story credible? 
		&  1& 90.9\%& 1 & 91.7\%& 1 & 91.7\%& 1 & 100\%\\

		\midrule
		6.-Does the text require your participation or cooperation to complete its form and meaning?
		&  1& 90.9\%& 0 &57.1\%& 1 & 85.7\%&  &\%\\
  		 
		\midrule
		7.-Does it propose a new vision of reality? 
		&  1& 90.9\%&  1 &66.7\%& 1 & 100\%& 1 &100\%\\

		\midrule
		8.-Does it propose a new vision of the genre it uses?  
		&  0& 61.11\%& 0 &100 \%& 0 & 100\%& 1 &100\%\\
	
		\midrule
		9.-Does it propose a new vision of the language itself? 
		&  1& 100\%& 0 &100\%& 0 & 100\%& 1 &100\%\\

		\midrule
		\multicolumn{9}{c}{Editorial/Commercial Quality}
         \\	
		\midrule
		10.-Does it remind you of another text or book you have read?
		&  1& 100 \%& 1 &100\%& 1 & 100\%& 1 &100\%\\

		\midrule
		11.-Would you like to read more texts like this? 
		&  1& 100 \%& 1 & 100 \%& 1 & 100\%& 1 &100\%\\
 	
		\midrule 
		12.-Would you recommend it? 
		&  1& 100 \%& 1 & 100\%& 1 & 100\%& 1 &100\%\\

		\midrule
		13.-Would you give it as a present? 
		&  1& 90.9\%& 1 & 100\%& 1 & 71.4\%& 1 &100\%\\
	
		\midrule

        Total by short story LLM generator 
         & 1
        & 92.42\%
        & 1
        &70.95\%
        & 1
        & 74.88\%
        & 1
        & 100
                 \%\\
        \bottomrule
	\end{tabularx}
\end{adjustwidth}
\end{table}

    \begin{table}[H]
	\small	
	\caption{Claude-opus-4-1-20250805 
 LLM responses to questions in Yes/No (Y/N) scale for all 48 short stories by source (The New Yorker, Claude V1.3, GPT-3.5-turbo and GPT-4). }
    \label{table:llms_ai_sht_yn_claude}
\begin{adjustwidth}{-\extralength}{0cm}

	\begin{tabularx}{\fulllength}{
			m{26em} c c  c  c  c  c c c
		} 
		\midrule
		\multicolumn{9}{c}{\textbf{Claude-opus-4-1-20250805 LLM Responses to Short Stories 1/0 (Y/N)}}\\
		\midrule 
        &\multicolumn{2}{c}{\textbf{The New Yorker}}
	& \multicolumn{2}{c}{\textbf{Claude}} & \multicolumn{2}{c}{\textbf{GPT-3.5-turbo}} & \multicolumn{2}{c}{\textbf{GPT-4}} \\
		\midrule
		\textbf{Question} & \textbf{Y/N} & \textbf{APR}& \textbf{Y/N} &\textbf{APR}& \textbf{Y/N} &\textbf{APR}& \textbf{Y/N}&\textbf{APR}\\
		\midrule
		\multicolumn{9}{c}{ Story Overview and Text Complexity}
        \\
		\midrule
		3.-Does it propose other interpretations, in~addition to the literal one? 
		&1 & 100\%& 1 & 87.5\%& 0 & 66.7\%& 1 & 100\%\\
		\midrule
		\multicolumn{9}{c}{Technical Assessment}
         \\
		\midrule
		5.-Is the story credible? 
			& 1 & 81.8\%& 1 & 100\%& 0 & 100\%& 1 & 100\%\\

		\midrule
		6.-Does the text require your participation or cooperation to complete its form and meaning?
		& 1 & 100\%& 0 & 83.33\%& 0 & 100\%& 1 & 87.5\%\\
  		 
		\midrule
		7.-Does it propose a new vision of reality? 
			& 1 & 100\%& 0 & 100\%& 1 & 100\%& 
            1 & 100\%\\

		\midrule
		8.-Does it propose a new vision of the genre it uses?  
			& 1& 100\%& 0 & 100\%& 0 & 100\%&  1&100\%\\
		\midrule
		9.-Does it propose a new vision of the language itself? 
			& 1& 100\%& 0& 100\%& 0 & 100\%& 1 & 100\%\\
		\midrule
		\multicolumn{9}{c}{Editorial/Commercial Quality}
         \\	
		\midrule
		10.-Does it remind you of another text or book you have read?
		& 1 & 100\%& 1 & 100\%& 1 & 100\%& 1 & 100\%\\

		\midrule
		11.-Would you like to read more texts like this? 
			& 1 & 100\%& 0& 80\%& 0 & 100\%& 1 & 100\%\\
 	
		\midrule 
		12.-Would you recommend it? 
			& 1 & 100\%& 1& 100\%&  0 & 100\%& 1 & 100\%\\

		\midrule
		13.-Would you give it as a present? 
		& 1 & 80\%& 1& 100\%& 0 & 100\%& 1 & 60\%\\
	
		\midrule

        Total by short story LLM generator
         &1
        & 98\%
        & 0
        & 68\%
        &0
        & 86.7\%
        & 1
        & 90.8 \%
        \\
        \bottomrule
	\end{tabularx}
\end{adjustwidth}
\end{table}

\section{Discussion}
\label{sec:discussion}

{Building on the results presented in Section~\ref{sec:results}, this discussion interprets the observed correlations and divergences between human and LLM-based evaluations, situating them within broader theoretical and practical frameworks. We reflect on the implications of these findings for literary studies, creative industries, and~automated assessment, emphasizing both the opportunities and the enduring challenges of integrating human and machine~judgment.

The intricate interplay of synergies and disparities between human and machine evaluation profoundly influences disciplines such as literary studies, educational assessment, and~the creative industries. Human evaluators contribute nuanced comprehension, emotional intelligence, and~deep contextual awareness—qualities that remain largely elusive to computational systems. This distinctly human faculty is indispensable in domains such as literary criticism, where the interpretation of thematic subtleties, affective resonance, and~sociocultural undertones demands a hermeneutic sensitivity that algorithms frequently fail to capture. A~human reader, for~instance, will discern the delicate layers of irony or the profound pathos of a narrative arc, dimensions that are often imperceptible to automated~analysis.

In contrast, machine evaluation excels in scalability, uniformity, and~rapid processing of extensive datasets. In~educational contexts, automated scoring engines can deliver instantaneous feedback, streamlining the assessment process and accommodating volumes that would otherwise overwhelm human graders; yet, this very efficiency harbors the danger of reductionism, potentially condensing richly textured creative expressions into reductive metrics. As such, the~central challenge is to harness quantitative precision without eroding the qualitative depth essential to a holistic appraisal of artistic and intellectual work. This balance becomes particularly critical in literary evaluation, where subtle interpretive variances often constitute the very essence of~creativity.

The evolving discourse on artificial intelligence must rigorously engage with these complementary and conflicting dynamics. As~AI capabilities advance, the~objective should be not to supplant human judgment, but to augment it. Hybrid evaluative frameworks in which machines perform initial processing and humans conduct deeper analysis could optimally merge computational speed with interpretative sophistication, thereby enriching outcomes across scholarly, pedagogical, and~creative domains. In~the context of the GrAImes and TTCW experiments, such a hybrid approach demonstrates that LLMs can serve as effective first-pass evaluators, highlighting promising or anomalous cases for human~interpretation.

This research with TTCW is the sole work employed to validate literary quality in microfictions and short stories. This validation is crucial, ensuring that the assessments are rigorous and reliable while providing a benchmark for future studies in the field. The~methodology used in this work has been meticulously validated by experts, ensuring that the evaluations are both accurate and~comprehensive.

This work pioneers the use of LLMs to judge the literary quality of text from an editorial perspective. This approach represents a significant advancement in the field, as~it leverages the capabilities of LLMs to provide nuanced and detailed evaluations. LLMs can perform adequate evaluation from an aesthetic and editorial point of view, complementing the TTCW test by considering the creative aspect. Additionally, this research begins to explore the importance of the variability of LLMs in responding to questions within the context of literary evaluation. This consideration of variability is essential for understanding the consistency and reliability of LLM-generated assessments, adding a layer of depth to the evaluation~process.

Furthermore, the~ethical dimensions of automated assessment demand critical scrutiny. Algorithmic bias, opacity in decision-making, and~the potential marginalization of human agency in creative practices represent significant concerns. Because AI models are trained on historical data, they risk perpetuating and even amplifying existing prejudices, thereby distorting evaluative outcomes and failing to capture the pluralism of human expression. A~rigorous ethical dialogue is imperative in order to develop AI systems that are not only efficient but also equitable, transparent, and~inclusive. Future frameworks should explicitly incorporate accountability mechanisms such as traceable evaluation logs or model explainability layers in order to safeguard interpretive integrity and fairness.}


\section{Conclusions}\label{sec:conclusion}

{This study set out to explore the extent to which LLM-based evaluators correlate with human judges in the assessment of creative writing. For this, we used two complementary frameworks: GrAImes and the Torrance Test of Creative Writing (TTCW). The~experiments provided quantitative and qualitative evidence that illuminates both the potential and the current limitations of AI in literary~evaluation.

Utilizing LLMs within the structured framework of the GrAImes evaluation protocol yields quantitatively measurable and methodologically replicable results, underscoring the potential of GrAImes as a scalable aid in the computational assessment of microfictions and short stories. The~empirical findings from this study indicate that while LLMs can effectively assist in the evaluation process by providing consistent large-volume judgments across numerous textual samples, they should be conceptualized as complementary instruments rather than as replacements for human judges. Integration of LLMs with the GrAImes protocol enabled a comprehensive multi-dimensional analysis of literary quality, as~the protocol's questions (translated into both Spanish and English to mitigate linguistic bias) systematically probe narrative complexity, technical proficiency, and~commercial viability in order to structure the models' evaluative~output.

It is essential to recognize that LLMs, despite their operational utility, cannot serve as definitive arbiters of literary quality due to inherent limitations in contextual understanding, emotional resonance, and~culturally-situated interpretation. Human raters, whether literary experts or enthusiasts, remain indispensable for assessing the nuanced, subjective, and~often implicit aspects of literary texts that exceed purely syntactic or semantic pattern recognition. These limitations are particularly evident in evaluative dimensions requiring deep intertextual knowledge, empathy, or~appreciation of stylistic innovation, areas where human judgment retains significant qualitative advantage. Nonetheless, the~consistency observed between human and LLM assessments in certain dimensions suggests that machine evaluation can provide valuable diagnostic and exploratory insights, particularly in large-scale or multilingual literary~corpora.

Nevertheless, rapid advancements in natural language processing such as improvements in transformer architectures, Reinforcement Learning from Human Feedback (RLHF), and~multimodal reasoning are progressively enhancing the evaluative and generative capabilities of these models. Such developments are enabling LLMs to not only assess but also generate narratives that exhibit increasingly sophisticated literary qualities, blurring the distinction between human- and machine-authored content in controlled contexts. This technological progression necessitates continued critical examination of the evaluative paradigms used to assess both human and synthetic literature. Future research should systematically investigate how these models' internal representations of style, emotion, and~narrative coherence correspond to human critical~categories.

Looking ahead, further research will expand the application of the GrAImes--LLM framework by incorporating a wider range of languages, diversifying literary genres and cultural contexts, and~engaging larger and more demographically varied pools of human respondents. Such expansion will facilitate robust cross-linguistic and cross-cultural validation of the protocol while enabling fine-grained analysis of model performance across different narrative traditions. This endeavor will not only refine our understanding of the synergies and disparities between human and machine evaluation but also contribute meaningfully to the evolving discourse on the role of artificial intelligence in literary studies, educational assessment, and~creative industries. Ultimately, the~convergence of computational precision and human interpretive depth points toward a new paradigm of collaborative evaluation, one that redefines creativity as an emergent property of dialogue between minds human and synthetic alike.}

		
		
		\vspace{6pt} 
		
		
		

		\authorcontributions
				{Conceptualization, G.A.M., R.M., H.C., and J.G.F.; methodology, R.M., H.C., and J.G.F.; software, G.A.M.; validation, G.A.M., R.M., and J.G.F.; formal analysis, R.M., H.C., and J.G.F.; investigation, G.A.M. and J.G.F.; resources, R.M. and J.G.F.; data curation, G.A.M. and J.G.F.; writing---original draft preparation, G.A.M.; writing---review and editing, R.M., H.C., J.G.F., and G.A.M.; visualization, G.A.M.; supervision, R.M., H.C., and J.G.F.; project administration, R.M.; funding acquisition, R.M. All authors have read and agreed to the published version of the manuscript.
		}
		
			
		\funding{This research was funded by ECOS NORD grant number~321105.}
		
		
		\informedconsent{Informed consent was obtained from all subjects involved in the study.}

		
		\dataavailability{The data presented in this study are openly available at: \url{
https://github.com/Manzanarez/mashup} (accessed on 1 December 2025).}

		\acknowledgments{
        During the preparation of this work, the authors used the Mistral Small 3 open source
 language model provided by Duck.ai
  and~the Language Tool  provided by Mozilla
    to improve the style of some sentences. The authors also used the large language models claude-sonnet-4-5-
20250929, Gemini-2.5-Flash-Preview-04-17, o3-2025-04-16, DeepSeek-V3-0324, GPT-4.1-
2025-04-14, gemini-2.5-pro, chatgpt-4o-latest-20250326, claude-opus-4-20250514, grok-4-0709, gemini-2.5-flash, gemini-2.5-flash-lite-preview-06-17-thinking, gpt-4.1-2025-04-14, o3-2025-04-16, grok-3-preview-02-24, deepseek-v3-0324, deepseek-r1-0528, claude-sonnet-4-20250514-thinking-32k, kimi-k2-0711-preview, hunyuan-turbos-20250416, qwen3-235b-a22b-no-thinking, mistral-medium-2505, claude-opus-4-1-20250805, chatgpt-4o-latest-20250326, and gpt-5-high for the purposes of the evaluation of microfictions and short stories applying GrAImes and to obtain the results used in this study. After~using these tools, the~authors reviewed and edited the content as needed, and take full responsibility for the content of the publication. We are grateful to the writers, literary experts, enthusiast readers, and~supporters who participated in this research, especially to Elisa A. Manzanarez, Amelia Olivia López, and Abel Vargas Guzmán.
    }
		
		\conflictsofinterest{The authors declare no conflicts of interest.
		
		
		
			
		}
		
		\appendixtitles{yes} 
		\appendixstart
		\appendix
        \section[Appendix Figures]{Figures and Tables of LLM Responses to GrAImes \linebreak  with Microfictions}
		\subsection[\appendixname~\thesubsection]{Krippendorff's $\alpha$}

        Krippendorffs's $\alpha$ dispersion graphics comparing every literary expert with all LLM responses to human-written microfictions.
\begin{figure}[H]

	\includegraphics[width=.495\linewidth]{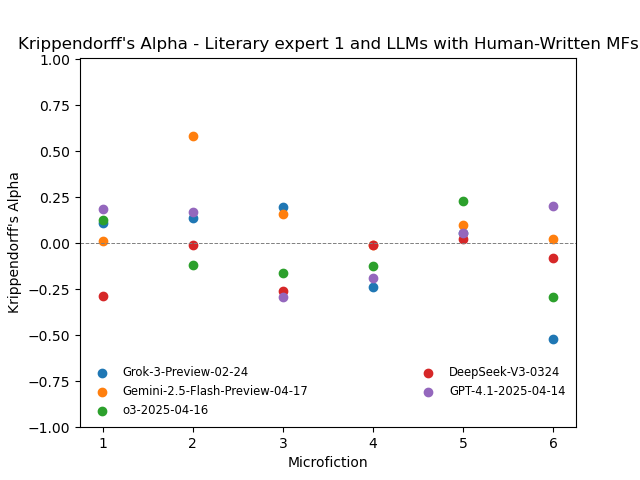}
    \includegraphics[width=.495\linewidth]{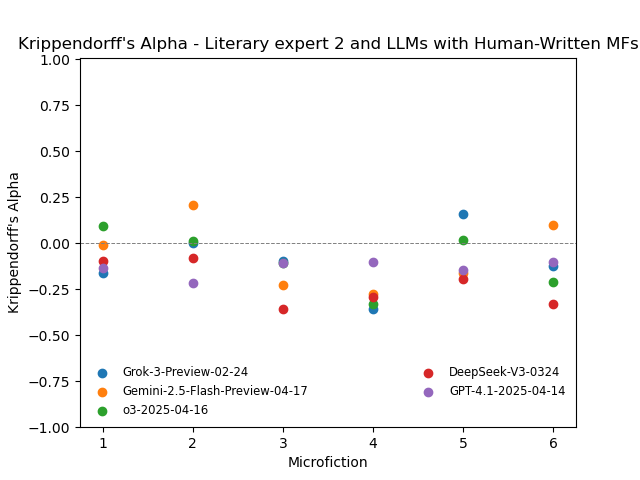}
    \includegraphics[width=.495\linewidth]{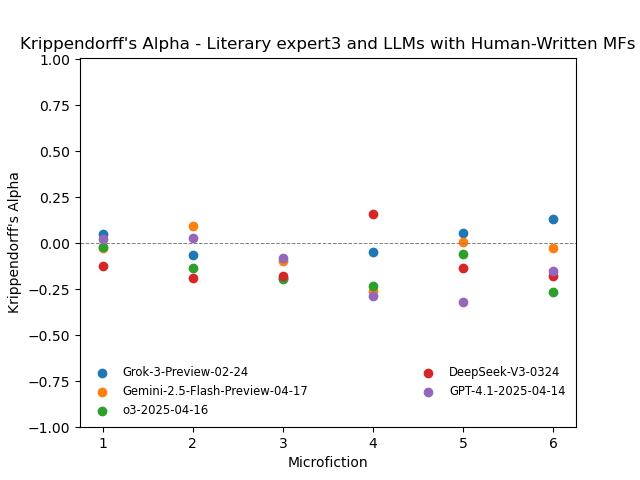}
    \includegraphics[width=.495\linewidth]{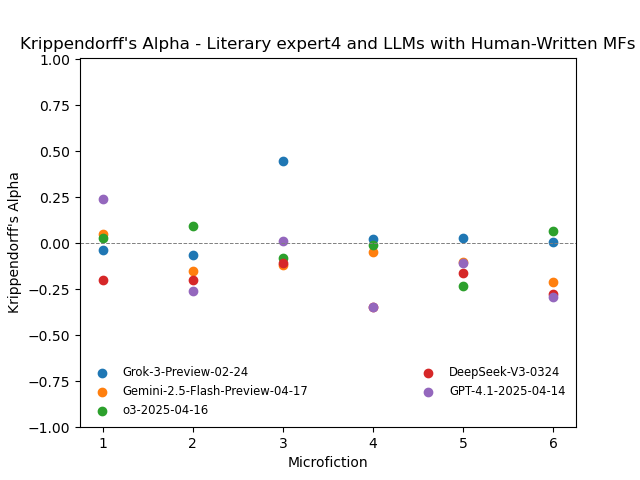}
    \includegraphics[width=.495\linewidth]{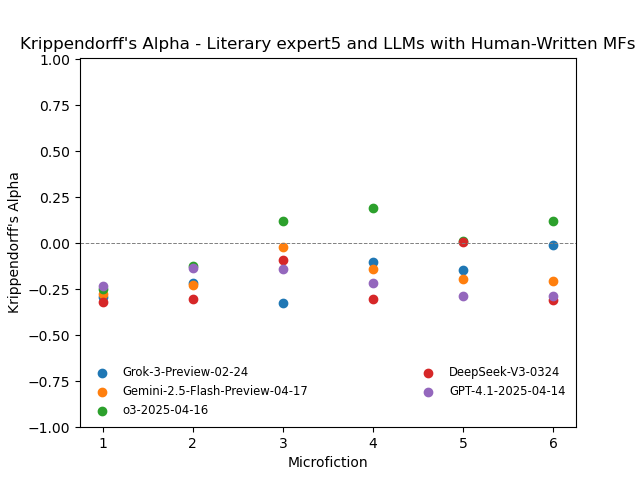}    \caption{Krippendorff's $\alpha$ for human experts and LLMs (as experts) with human-written~microfictions.}
	\label{fig:Krip_alpha_le1_llms_hw}
\end{figure}


\begin{table}[H]
	\caption{Krippendorff's $\alpha$ and Confidence Interval (CI) of LLMs and literary expert 2 using GrAImes with human-written microfictions. LLM 1 = Grok-3-Preview-02-24, LLM 2 = Gemini-2.5-Flash-Preview-04-17, LLM 3 = o3-2025-04-16, LLM 4 = DeepSeek-V3-0324, LLM 5 = GPT-4.1-2025-04-14}.
	\label{table:Krippendorf_alpha_le1_2_llms_hw_mfs}
	\scriptsize
\begin{adjustwidth}{-\extralength}{0cm}

	\begin{tabular}{p{.25cm} p{.55cm} p{1.65cm} 
    p{.55cm} p{1.6cm} 
    p{.55cm} p{1.55cm} 
    p{.55cm} p{1.55cm} 
    p{.55cm} p{1.6cm} 
    p{.55cm} p{1.5cm} }
        \midrule
		\multicolumn{13}{c}{\textbf{Krippendorff's \boldmath{$\alpha$} and Confidence Interval (CI) for Literary Expert 2}}  \\
		\midrule
		\textbf{LLM} & \textbf{MF1} & &\textbf{MF2} && \textbf{MF3} &&\textbf{MF4} &&\textbf{MF5} && \textbf{MF6} \\ 
		\midrule	
		 & \boldmath{$\alpha$}&\textbf{CI}& \boldmath{$\alpha$} & \textbf{CI}& \boldmath{$\alpha$} & \textbf{CI}& \boldmath{$\alpha$} & \textbf{CI}& \boldmath{$\alpha$}  & \textbf{CI}& \boldmath{$\alpha$} & \textbf{CI} \\ 
		\midrule	
        1 &$-$0.163 &  (0.088, 0.388)& 0.0 &  ($-$0.046, 0.282)& $-$0.094 &  (0.004, 0.460)& $-$0.357 &  (0.190, 0.693))& 0.159 &  ($-$0.046, 0.262)& $-$0.125 &  (0.070, 0.386) \\ \midrule
        2 &$-$0.008 &  ($-$0.017, 0.344))& 0.208 &  ($-$0.039, 0.188)& $-$0.226 &  (0.033, 0.404)& $-$0.275 &  (0.153, 0.578)& $-$0.163 &  (0.015, 0.296)& 0.101&   (0.018, 0.311)\\ \midrule
        3 &0.095&  (0.003, 0.331)& 0.066 &  ($-$0.046, 0.215)& 0.015 &  (0.034, 0.531)& $-$0.329 &  (0.134, 0.502)& 0.017 &  ($-$0.046, 0.252) &$-$0.213&  (0.156, 0.592)\\ \midrule
        4 &0.027 &  (0.035, 0.480)& $-$0.081 &  (0.051, 0.549)& $-$0.094&  (0.056, 0.476)& $-$0.293 &  (0.153, 0.578)& $-$0.196 &  ($-$0.002, 0.337)& $-$0.329&  (0.144, 0.571)\\ \midrule
        5 & $-$0.132 &  (0.066, 0.551)& $-$0.216 &  ($-$0.017, 0.648)& $-$0.109 &  (0.081, 0.561)& $-$0.103 &  (0.100, 0.488)& $-$0.147 &  ($-$0.046, 0.262)& $-$0.101&   (0.054, 0.464)\\ \midrule

	\end{tabular}	
\end{adjustwidth}
\end{table}
\unskip

\begin{table}[H]
	\captionof{table}{Krippendorff's $\alpha$ and Confidence Interval (CI) of LLMs and literary expert 3 using GrAImes with human-written microfictions. LLM 1 = Grok-3-Preview-02-24, LLM 2 = Gemini-2.5-Flash-Preview-04-17, 
LLM 3 = o3-2025-04-16, LLM 4 = DeepSeek-V3-0324, LLM 5 = GPT-4.1-2025-04-14.}
	\label{table:Krippendorf_alpha_le1_3_llms_hw_mfs}
	\scriptsize
\begin{adjustwidth}{-\extralength}{0cm}

	\begin{tabular}{p{.25cm} p{.55cm} p{1.6cm} 
    p{.55cm} p{1.6cm} 
    p{.55cm} p{1.55cm} 
    p{.55cm} p{1.6cm} 
    p{.55cm} p{1.6cm} 
    p{.55cm} p{1.5cm} }
		\midrule
		\multicolumn{12}{c}{\textbf{Krippendorff's \boldmath{$\alpha$} and Confidence Interval (CI) for Literary Expert 3}}  \\
        		\midrule	
		\textbf{LLM} & \textbf{MF1} & &\textbf{MF2} && \textbf{MF3} &&\textbf{MF4} &&\textbf{MF5} && \textbf{MF6} \\ 
		\midrule	
		 & \boldmath{$\alpha$}&\textbf{CI}& \boldmath{$\alpha$} & \textbf{CI}& \boldmath{$\alpha$} & \textbf{CI}& \boldmath{$\alpha$} & \textbf{CI}& \boldmath{$\alpha$}  & \textbf{CI}& \boldmath{$\alpha$} & \textbf{CI} \\ 
		\midrule 	
        1 & 0.050 &  (0.002, 0.287)& $-$0.063 &  (0.017, 0.286)& $-$0.188 &  (0.023, 0.488)& $-$0.047 &  (0.013, 0.332)& 0.057 &  (0.004, 0.356)& 0.131 &  (0.016, 0.571) \\ \midrule
        2 & $-$0.027 &  (0.023, 0.373)& 0.095 &  ($-$0.029, 0.235)& $-$0.096 &  (0.011, 0.397)& $-$0.257 &  ($-$0.022, 0.273)& 0.007 &  ($-$0.033, 0.181)& $-$0.027 &  (0.001, 0.480)\\ \midrule
        3 & $-$0.020 &  ($-$0.013, 0.268)& $-$0.132 &  ($-$0.003, 0.296)& $-$0.196 &  (0.079, 0.578)& $-$0.230 &  ($-$0.025, 0.296)& $-$0.056 &  (0.000, 0.286)& $-$0.267&   (0.146, 0.578))\\ \midrule
        4 & $-$0.126 &  ($-$0.032, 0.410)& $-$0.188 &  ($-$0.015, 0.385)& $-$0.179 &  (0.084, 0.485)& 0.162 &  ($-$0.022, 0.273)& $-$0.132 &  ($-$0.026, 0.263)& $-$0.178 &  (0.073, 0.665)\\ \midrule
        5 & 0.022 &  ($-$0.031, 0.344)& 0.029 &  ($-$0.025, 0.336)& $-$0.078 &  (0.084, 0.565)& $-$0.284 &  ($-$0.026, 0.251)& $-$0.319 &  (0.004, 0.356)& $-$0.148&  (0.091, 0.578)\\ \midrule

	\end{tabular}
\end{adjustwidth}	
\end{table}
\unskip

\begin{table}[H]
	\captionof{table}{Krippendorff's $\alpha$ and Confidence Interval (CI) of LLMs and literary expert 4 using GrAImes with human-written microfictions. LLM 1 = Grok-3-Preview-02-24, LLM 2 = Gemini-2.5-Flash-Preview-04-17, 
LLM 3 = o3-2025-04-16, 4 = DeepSeek-V3-0324, 
LLM 5 = GPT-4.1-2025-04-14}
	\label{table:Krippendorf_alpha_le1_4_llms_hw_mfs}
	\scriptsize
\begin{adjustwidth}{-\extralength}{0cm}

	\begin{tabular}{p{.25cm} p{.54cm} p{1.6cm} 
    p{.54cm} p{1.6cm} 
    p{.54cm} p{1.6cm} 
    p{.54cm} p{1.5cm} 
    p{.54cm} p{1.6cm} 
    p{.54cm} p{1.6cm} }
		\midrule
		\multicolumn{12}{c}{\textbf{Krippendorff's \boldmath{$\alpha$} and Confidence Interval (CI) for Literary Expert 4}}  \\
		\midrule
		\textbf{LLM} & \textbf{MF1} & &\textbf{MF2} && \textbf{MF3} &&\textbf{MF4} &&\textbf{MF5} && \textbf{MF6} \\ 
		\midrule	
		 & \boldmath{$\alpha$}&\textbf{CI}& \boldmath{$\alpha$} & \textbf{CI}& \boldmath{$\alpha$} & \textbf{CI}& \boldmath{$\alpha$} & \textbf{CI}& \boldmath{$\alpha$}  & \textbf{CI}& \boldmath{$\alpha$} & \textbf{CI} \\ 
		\midrule 		
        1 & $-$0.039 &  ($-$0.028, 0.214)& $-$0.063 &  (0.017, 0.297)& 0.445 &  ($-$0.048, 0.117)& 0.026 &  (0.016, 0.648)& 0.029 &  ($-$0.027, 0.273)& 0.007&  (0.010, 0.336) \\ \midrule
        2 & 0.050 &  (0.008, 0.408)& $-$0.148 &  ($-$0.016, 0.307)& $-$0.118 &  ($-$0.027, 0.214)& $-$0.047 &  (0.029, 0.634)& $-$0.103 &  ($-$0.019, 0.221)& $-$0.210&  (0.032, 0.316)\\ \midrule
        3& 0.029 &  ($-$0.039, 0.245)& 0.095  & ($-$0.014, 0.253)& $-$0.08 &  ($-$0.031, 0.357)& $-$0.008 &  (0.026, 0.485)& $-$0.230 &  ($-$0.020, 0.331)& 0.066&  ($-$0.013, 0.372) \\ \midrule
        4 & $-$0.197 &  ($-$0.035, 0.324)& $-$0.197 &  ($-$0.028, 0.438)& $-$0.109 &  ($-$0.013, 0.257)& $-$0.346 &  (0.029, 0.634)& $-$0.163 &  ($-$0.033, 0.256)& $-$0.275&  (0.144, 0.500)\\ \midrule
        5 & 0.240 &  ($-$0.047, 0.239)& $-$0.257 &  ($-$0.025, 0.383)& 0.015 &  (0.008, 0.332)& $-$0.348 &  (0.042, 0.468)& $-$0.109 &  ($-$0.027, 0.273)& $-$0.293&  (0.022, 0.476)\\ \midrule

	\end{tabular}
\end{adjustwidth}	
\end{table}
\vspace{-6pt}

\begin{table}[H]
	\captionof{table}{Krippendorff's $\alpha$ and Confidence Interval (CI) of LLMs and literary expert 5 using GrAImes with human-written microfictions. LLM 1 = Grok-3-Preview-02-24, LLM 2 = Gemini-2.5-Flash-Preview-04-17, 
LLM 3 = o3-2025-04-16, LLM 4 = DeepSeek-V3-0324, LLM 5 = GPT-4.1-2025-04-14.}
	\label{table:Krippendorf_alpha_le1_5_llms_hw_mfs}
	\scriptsize
\begin{adjustwidth}{-\extralength}{0cm}

	\begin{tabular}{p{.25cm} p{.55cm} p{1.5cm} 
    p{.55cm} p{1.5cm} 
    p{.55cm} p{1.6cm} 
    p{.55cm} p{1.6cm} 
    p{.55cm} p{1.6cm} 
    p{.55cm} p{1.62cm} }
		\midrule
		\multicolumn{12}{c}{\textbf{Krippendorff's \boldmath{$\alpha$} and Confidence Interval (CI) for Literary Expert 5}}  \\
		\midrule
		\textbf{LLM} & \textbf{MF1} & &\textbf{MF2} && \textbf{MF3} &&\textbf{MF4} &&\textbf{MF5} && \textbf{MF6} \\ 
		\midrule	
		 & \boldmath{$\alpha$}&\textbf{CI}& \boldmath{$\alpha$} & \textbf{CI}& \boldmath{$\alpha$} & \textbf{CI}& \boldmath{$\alpha$} & \textbf{CI}& \boldmath{$\alpha$}  & \textbf{CI}& \boldmath{$\alpha$} & \textbf{CI} \\ 
		\midrule	
        1 &$-$0.293&  (0.144, 0.513)& $-$0.218 &  (0.225, 0.430)& $-$0.326 &  ($-$0.004, 0.434)& $-$0.101 &  (0.060, 0.500)& $-$0.143 &  ($-$0.013, 0.368)& $-$0.007&  (0.054, 0.341)\\ \midrule
        2 & $-$0.275 &  (0.134, 0.592)& $-$0.226 &  (0.119, 0.430)& $-$0.020 &  (0.026, 0.372)& $-$0.140 &  (0.022, 0.500)& $-$0.195 &  (0.060, 0.296)&  $-$0.203&  (0.083, 0.388)\\ \midrule
        3 & $-$0.250 &  (0.111, 0.464)& $-$0.125 &  (0.077, 0.430)& 0.120 &  ($-$0.025, 0.209)& 0.191 &  ($-$0.005, 0.306)& 0.015 &  ($-$0.019, 0.350)& 0.120&  ($-$0.026, 0.369)\\ \midrule
        4 & $-$0.319&  (0.095, 0.675)& $-$0.301 &  (0.193, 0.578)& $-$0.090 &  (0.022, 0.340)& $-$0.301 &  (0.022, 0.500)& 0.007 &  (0.028, 0.306)& $-$0.310&  (0.167, 0.551)\\ \midrule
        5 & $-$0.230&  (0.070, 0.551)& $-$0.132 &  (0.134, 0.538)& $-$0.169&  (0.078, 0.549)& $-$0.218 &  (0.052, 0.325)& $-$0.286 &  ($-$0.013, 0.368)& $-$0.284&  (0.083, 0.480) \\ \midrule
	\end{tabular}
\end{adjustwidth}	
\end{table}
\vspace{-6pt}

\begin{figure}[H]
\vspace{-6pt}
   \hspace*{-6pt} \includegraphics[width=0.5\linewidth]{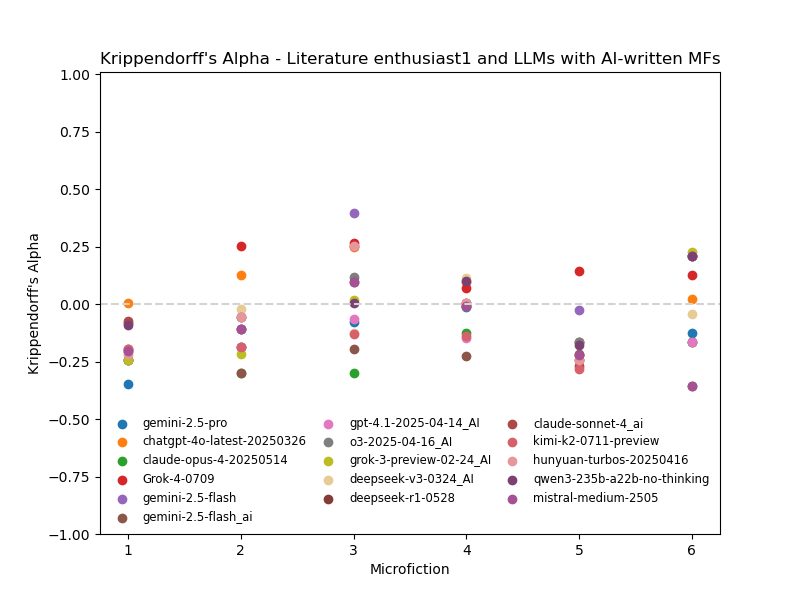}
    \includegraphics[width=0.5\linewidth]{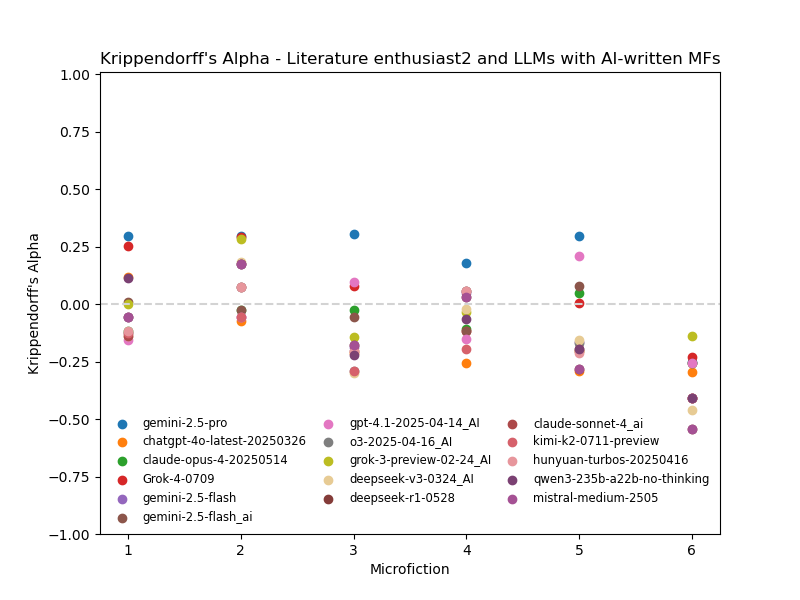}\\
   \hspace*{-6pt}  \includegraphics[width=0.5\linewidth]{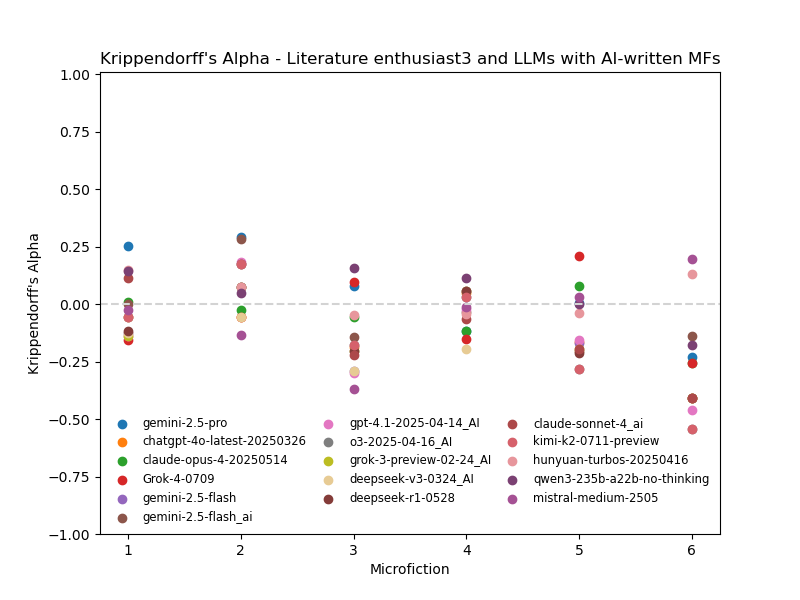}
    \includegraphics[width=0.5\linewidth]{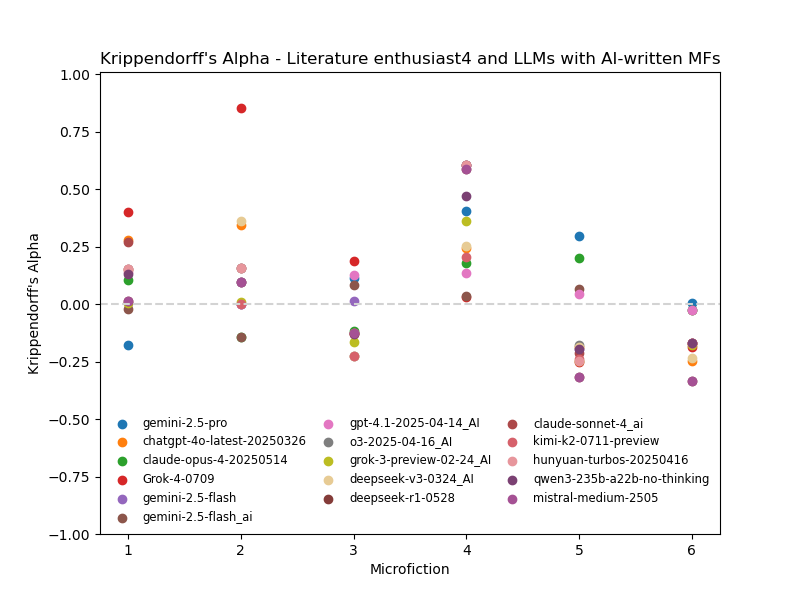}\\
   \hspace*{-6pt}  \includegraphics[width=0.5\linewidth]{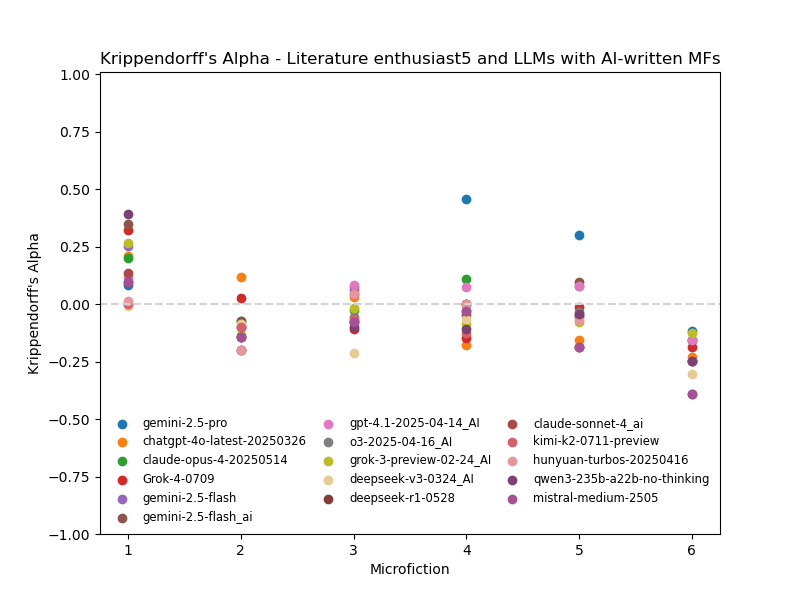}
    \includegraphics[width=0.5\linewidth]{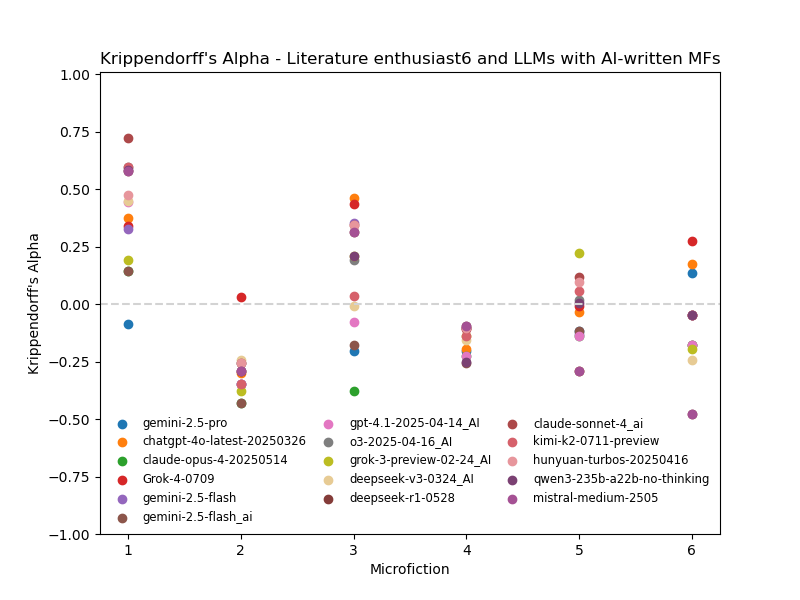}    \caption{Krippendorff's $\alpha$ for 16 literary enthusiasts (16 LLMs).}
    \label{fig:kripendorff_lit_enth16_llms16}

\end{figure}
\vspace{-6pt}

\begin{table}[H]
\caption{Krippendorff's $\alpha$ for literary enthusiast 2 (16 LLMs).}
\label{table:Krippendorf_alpha_le2_llms_ai_mfs}
	\small
	\begin{tabularx}{\textwidth}{l C C C C C C}
		\toprule
		\multicolumn{6}{c}{\textbf{Krippendorff's \boldmath{$\alpha$} Literary Enthusiast 2---LLMs (16)}}  \\
		\midrule
		\textbf{LLM} &\textbf{ MF1} & \textbf{MF2} & \textbf{MF3} &\textbf{MF4} &\textbf{MF5} & \textbf{MF6} \\ 
		\midrule		
gemini-2.5-pro& 0.296	&0.296&	0.305&	0.181&	0.296&	$-$0.25\\
        \midrule
chatgpt-4o-latest-20250326 & 0.116&	$-$0.073&	$-$0.213&	$-$0.258&	$-$0.293&	$-$0.295\\ 
        \midrule
claude-opus-4-20250514 &$-$0.118&	$-$0.027&	$-$0.027&	$-$0.109&	0.05&	$-$0.258\\
        \midrule
grok-4-0709 &0.252	&0.29&	0.081&	$-$0.118&	0.007&$-$0.23\\ 
        \midrule
gemini-2.5-flash &$-$0.056&	$-$0.056&	$-$0.188&	0.05	&$-$0.163&	$-$0.407\\ 
        \midrule
gemini-2.5-flash\_ai  &0.01&	$-$0.027&	$-$0.056&	$-$0.118&	0.081&	$-$0.258\\ 
    \midrule
gpt-4.1-2025-04-14\_AI &$-$0.157&	$-$0.056&	0.095&	$-$0.152&	0.208&	$-$0.258
\\ 
        \midrule
o3-2025-04-16\_AI &$-$0.127&	0.073&	$-$0.293&	0.057&	$-$0.171&	$-$0.407\\ 
        \midrule
grok-3-preview-02-24\_AI &  0&	0.283&	$-$0.143&	$-$0.034&	$-$0.204&	$-$0.14\\
\bottomrule
\end{tabularx} 
\end{table}

\begin{table}[H]\ContinuedFloat
\caption{\textit{Cont.}}
	\small
	\begin{tabularx}{\textwidth}{l C C C C C C}
		\toprule		
		\multicolumn{6}{c}{\textbf{Krippendorff's \boldmath{$\alpha$} for Literary Enthusiast 2---LLMs (16)}}  \\
		\midrule
		\textbf{LLM} &\textbf{ MF1} & \textbf{MF2} & \textbf{MF3} &\textbf{MF4} &\textbf{MF5} & \textbf{MF6} \\ 
        \midrule
deepseek-v3-0324\_AI & $-$0.137&	0.183&	$-$0.301&	$-$0.02&	$-$0.155&	$-$0.462\\
        \midrule
deepseek-r1-0528 &  $-$0.056&	0.174&	$-$0.179&	0.029&	$-$0.284&	$-$0.545\\
        \midrule
claude-sonnet-4\_ai & $-$0.137&	0.073&	$-$0.204&	0.057&	$-$0.196&	$-$0.407\\
        \midrule
kimi-k2-0711-preview & $-$0.127&	$-$0.056&	$-$0.293&	$-$0.196&	$-$0.204&	$-$0.407\\
        \midrule
hunyuan-turbos-20250416   & $-$0.118&	0.073&	$-$0.204&	0.057&	$-$0.213&	$-$0.407\\
        \midrule
qwen3-235b-a22b-no-thinking  & 0.112&	0.174&	$-$0.221&	$-$0.063&	$-$0.196&	$-$0.407\\
        \midrule
                Mistral-medium-2505& $-$0.056&	0.174&	$-$0.179&	0.029&	$-$0.284&	$-$0.545\\
\bottomrule
	\end{tabularx}	
\end{table}
\unskip

\begin{figure}[H]

   \hspace*{-6pt} \includegraphics[width=0.5\linewidth]{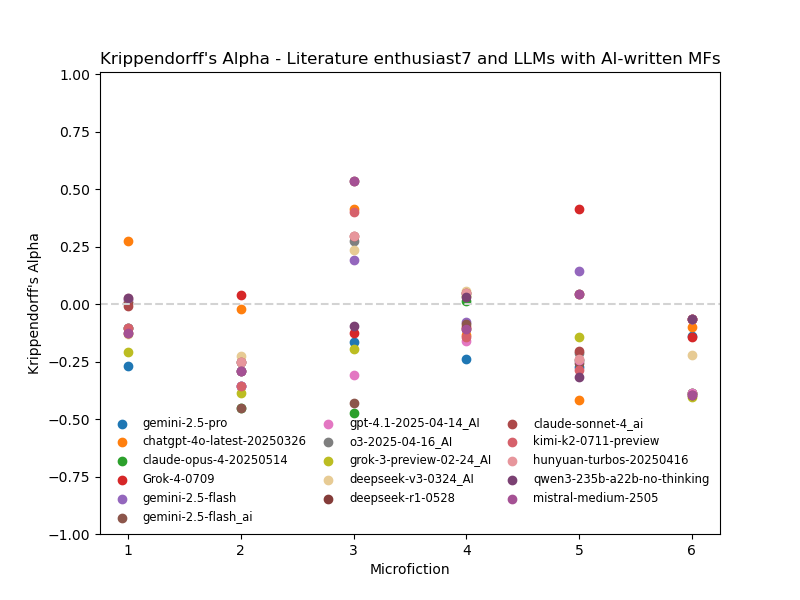}
    \includegraphics[width=0.5\linewidth]{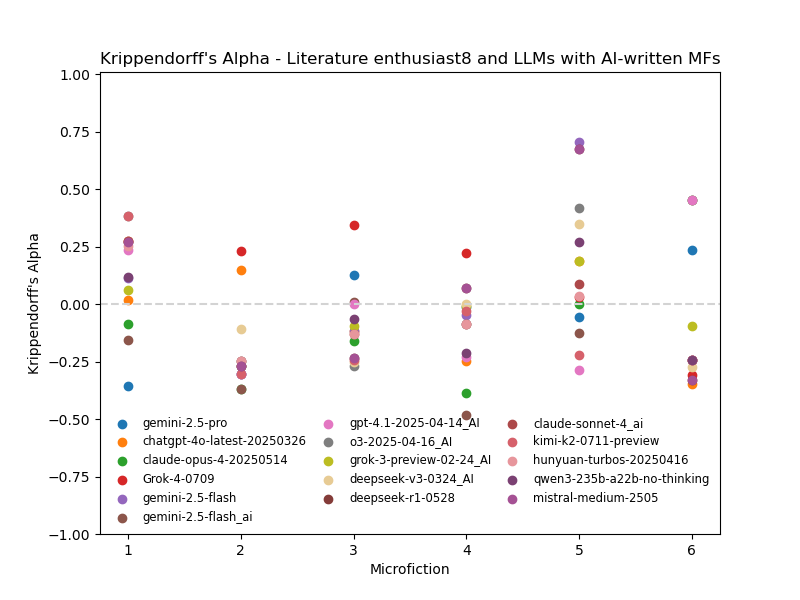}\\
   \hspace*{-6pt} \includegraphics[width=0.5\linewidth]{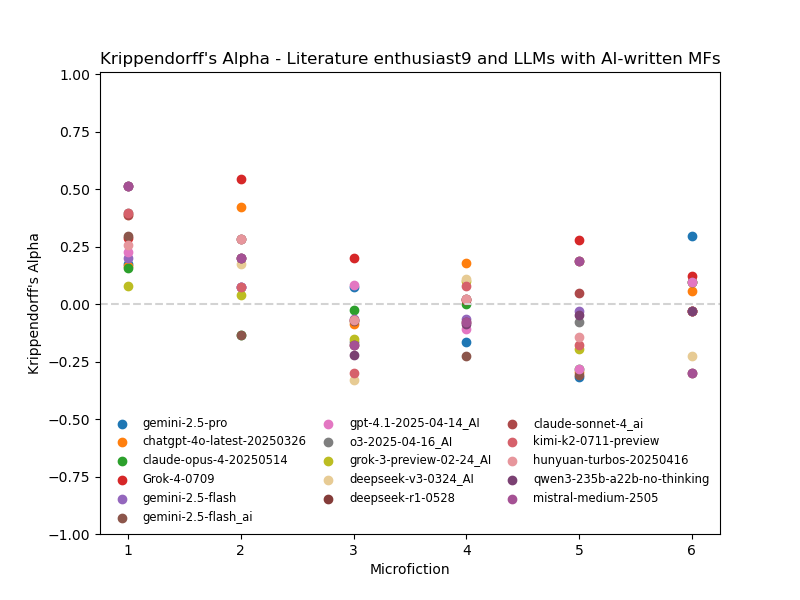}
    \includegraphics[width=0.5\linewidth]{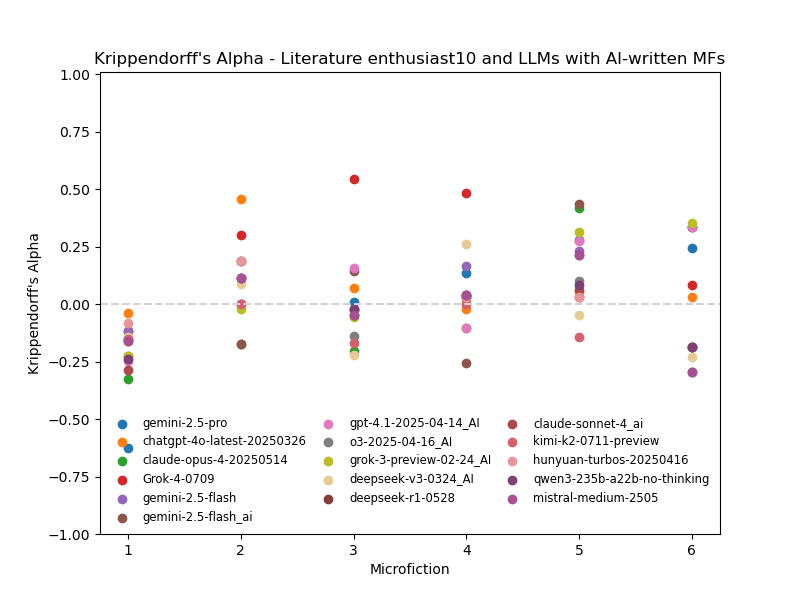}
    
    \caption{Krippendorff's $\alpha$ for literary enthusiasts (16 LLMs).}
         \label{fig:kripendorff_lit_enth16_llms16_2}
\end{figure}
\vspace{-6pt}

\begin{table}[H]
\caption{Krippendorff's $\alpha$ for literary enthusiast 3 (16 LLMs).}
\label{table:Krippendorf_alpha_le3_llms_ai_mfs}
	\small
		\begin{tabularx}{\textwidth}{l C C C C C C}
		\toprule
		\multicolumn{6}{c}{\textbf{Krippendorff's \boldmath{$\alpha$} Literary enthusiast 3---LLMs (16)}}  \\
		\midrule
		\textbf{LLM} &\textbf{ MF1} & \textbf{MF2} & \textbf{MF3} &\textbf{MF4} &\textbf{MF5} & \textbf{MF6} \\ 
		\midrule		
gemini-2.5-pro&0.252&	0.29&	0.081&	$-$0.118&	0.007&	$-$0.23\\
        \midrule
chatgpt-4o-latest-20250326 &
$-$0.056&	$-$0.056&	$-$0.188&	0.05&	$-$0.163&	$-$0.407\\
        \midrule
claude-opus-4-20250514 & 
0.01&	$-$0.027&	$-$0.056&	$-$0.118&	0.081&	$-$0.258\\
        \midrule
grok-4-0709 & 
$-$0.157&	$-$0.056&	0.095&	$-$0.152&	0.208&	$-$0.258\\
        \midrule
gemini-2.5-flash &
$-$0.127&	0.073&	$-$0.293&	0.057&	$-$0.171&	$-$0.407\\
        \midrule
gemini-2.5-flash\_ai  &0&	0.283&	$-$0.143&	$-$0.034&	$-$0.204&	$-$0.14\\
        \midrule
gpt-4.1-2025-04-14\_AI & $-$0.137&	0.183&	$-$0.301&	$-$0.02&	$-$0.155&$-$0.462\\
        \midrule
o3-2025-04-16\_AI & $-$0.056&	0.174&	$-$0.179&	0.029&	$-$0.284&	$-$0.545\\
\bottomrule
\end{tabularx} 
\end{table}

\begin{table}[H]\ContinuedFloat
\caption{\textit{Cont.}}
	\small
		\begin{tabularx}{\textwidth}{l C C C C C C}
		\toprule
		\multicolumn{6}{c}{\textbf{Krippendorff's \boldmath{$\alpha$} Literary enthusiast 3---LLMs (16)}}  \\
		\midrule
		\textbf{LLM} &\textbf{ MF1} & \textbf{MF2} & \textbf{MF3} &\textbf{MF4} &\textbf{MF5} & \textbf{MF6} \\ 
		\midrule	
grok-3-preview-02-24\_AI & $-$0.137&	0.073&	$-$0.204&	0.057&	$-$0.196&	$-$0.407\\
        \midrule
deepseek-v3-0324\_AI & $-$0.127&	$-$0.056&	$-$0.293&	$-$0.196&	$-$0.204&	$-$0.407\\
        \midrule
deepseek-r1-0528 & $-$0.118&	0.073&	$-$0.204&	0.057&	$-$0.213&	$-$0.407\\
        \midrule
claude-sonnet-4\_ai &0.112&	0.174&	$-$0.221&	$-$0.063&	$-$0.196&	$-$0.407\\

        \midrule
kimi-k2-0711-preview & $-$0.056&	0.174&	$-$0.179&	0.029&	$-$0.284&	$-$0.545\\
        \midrule
hunyuan-turbos-20250416   & 0.149&	0.073&	$-$0.047&	$-$0.041&	$-$0.039&	0.131\\
        \midrule
qwen3-235b-a22b-no-thinking  &0.143&	0.05&	0.156&	0.113&	0&	$-$0.179\\ 
        \midrule
                Mistral-medium-2505& $-$0.027&	$-$0.134&	$-$0.371&	$-$0.013&	0.029&	0.197\\
 \bottomrule

    \end{tabularx}	
\end{table}
\unskip

\begin{table}[H]
\caption{Krippendorff's $\alpha$ for literary enthusiast 4 (16 LLMs).}
\label{table:Krippendorf_alpha_le4_llms_ai_mfs}
		\small
		\begin{tabularx}{\textwidth}{l C C C C C C}
		\toprule
		\multicolumn{6}{c}{\textbf{Krippendorff's \boldmath{$\alpha$} Literary enthusiast 4 - LLMs (16)}}  \\
		\midrule
		\textbf{LLM} &\textbf{ MF1} & \textbf{MF2} & \textbf{MF3} &\textbf{MF4} &\textbf{MF5} & \textbf{MF6} \\ 
		\midrule		
gemini-2.5-pro&$-$0.178&	0&	0.112&	0.406&	0.296&	0.007\\
        \midrule
chatgpt-4o-latest-20250326 &0.278&	0.345&	$-$0.132&	0.245&	$-$0.188&	$-$0.248\\
        \midrule
claude-opus-4-20250514 &0.107&	$-$0.145&	$-$0.118&	0.18&	0.202&	$-$0.027\\
        \midrule
grok-4-0709 &0.399&	0.855&	0.186&	0.032&	$-$0.25&	$-$0.188\\
        \midrule
gemini-2.5-flash &0.007&	0&	0.013&	0.604&	$-$0.196&	$-$0.169\\
        \midrule
gemini-2.5-flash\_ai  &$-$0.02&	$-$0.145&	0.084&	0.036&	0.066&	$-$0.027\\
        \midrule
gpt-4.1-2025-04-14\_AI &0.142&	0&	0.128&	0.136&	0.042&	$-$0.027\\
        \midrule
o3-2025-04-16\_AI &0.153&	0.159&	$-$0.226&	0.607&	$-$0.179&	$-$0.169\\
        \midrule
grok-3-preview-02-24\_AI &0.007&	0.01&	$-$0.163&	0.362&	$-$0.248&	$-$0.179\\
        \midrule
deepseek-v3-0324\_AI &0.147&	0.361&	$-$0.226&	0.255&	$-$0.188&	$-$0.236\\
        \midrule
deepseek-r1-0528 &0.013&	0.095&	$-$0.125&	0.59&	$-$0.319&	$-$0.333\\
        \midrule
claude-sonnet-4\_ai &0.269&	0.159&	$-$0.125&	0.607&	$-$0.213&	$-$0.169\\
        \midrule
kimi-k2-0711-preview &0.153&	0&	$-$0.226&	0.203&	$-$0.239&	$-$0.169\\
        \midrule
hunyuan-turbos-20250416  &0.153&	0.159&	$-$0.125&	0.607&	$-$0.248&	$-$0.169\\
        \midrule
qwen3-235b-a22b-no-thinking  & 0.131&	0.095&	$-$0.132&	0.469&	$-$0.196&	$-$0.169\\

        \midrule
                Mistral-medium-&0.013&	0.095&	$-$0.125&	0.59&	$-$0.319&	$-$0.333\\
 \bottomrule

    \end{tabularx}	
\end{table}
\unskip

\begin{table}[H]
\caption{Krippendorff's $\alpha$ for literary enthusiast 5 (16 LLMs).}
\label{table:Krippendorf_alpha_le5_llms_ai_mfs}
		\small
		\begin{tabularx}{\textwidth}{l C C C C C C}
		\toprule
		\multicolumn{6}{c}{\textbf{Krippendorff's \boldmath{$\alpha$} Literary enthusiast 5 - LLMs (16)}}  \\
		\midrule
		\textbf{LLM} &\textbf{ MF1} & \textbf{MF2} & \textbf{MF3} &\textbf{MF4} &\textbf{MF5} & \textbf{MF6} \\ 
		\midrule		
gemini-2.5-pro&0.084&	$-$0.101&	0.073&	0.457&	0.303&	$-$0.118\\
        \midrule
chatgpt-4o-latest-20250326 &0.208&	0.116&	0.029&	$-$0.179&	$-$0.155&	$-$0.23\\
        \midrule
claude-opus-4-20250514 & 0.202&	$-$0.075&	$-$0.027&	0.109&	0.081&	$-$0.155\\
        \midrule
grok-4-0709 &0.321&	0.026&	$-$0.108&	$-$0.148&	$-$0.013&	$-$0.188\\
        \midrule
gemini-2.5-flash &0.252&	$-$0.101&	0.063&	0&	$-$0.056&	$-$0.248\\
        \midrule
gemini-2.5-flash\_ai  &0.35&	$-$0.075&	$-$0.075&	$-$0.047&	0.095&	$-$0.155\\
        \midrule
gpt-4.1-2025-04-14\_AI &0.116&	$-$0.101&	0.084&	0.073&	0.081&	$-$0.155\\
\bottomrule
\end{tabularx} 
\end{table}

\begin{table}[H]\ContinuedFloat
\caption{\textit{Cont.}}
		\small
		\begin{tabularx}{\textwidth}{l C C C C C C}
		\toprule
		\multicolumn{6}{c}{\textbf{Krippendorff's \boldmath{$\alpha$} Literary enthusiast 5 - LLMs (16)}}  \\
		\midrule
		\textbf{LLM} &\textbf{ MF1} & \textbf{MF2} & \textbf{MF3} &\textbf{MF4} &\textbf{MF5} & \textbf{MF6} \\ 
        \midrule
o3-2025-04-16\_AI &0&	$-$0.2&	$-$0.056&	0&	$-$0.034&	$-$0.248\\
        \midrule
grok-3-preview-02-24\_AI &0.264&	$-$0.134&	$-$0.015&	$-$0.086&	$-$0.078&	$-$0.125\\
        \midrule
deepseek-v3-0324\_AI &$-$0.008&	$-$0.086&	$-$0.213&	$-$0.063&	$-$0.188&	$-$0.305\\
        \midrule
deepseek-r1-0528 &0.095&	$-$0.145&	$-$0.078&	$-$0.031&	$-$0.188&	$-$0.39\\
        \midrule
claude-sonnet-4\_ai &0.136&	$-$0.2&	0.043&	0&	$-$0.048&	$-$0.248\\
        \midrule
kimi-k2-0711-preview &0&	$-$0.101&	$-$0.063&	$-$0.126&	$-$0.188&	$-$0.248\\
        \midrule
hunyuan-turbos-20250416  &0.015&	$-$0.2&	0.043&	0&	$-$0.07&	$-$0.248\\
        \midrule
qwen3-235b-a22b-no-thinking  &0.392&	$-$0.145&	$-$0.101&	$-$0.109&	$-$0.041&	$-$0.248\\ 
        \midrule
                Mistral-medium-&0.095&	$-$0.145&	$-$0.078&	$-$0.031&	$-$0.188&	$-$0.39\\
 \bottomrule
    \end{tabularx}	
\end{table}

\vspace{-6pt}

\begin{table}[H]
		\small
\caption{GrAImes expert evaluation: LLMs used as judges with the expert prompt and responses compared against human experts.}
    \label{table:llms_exps_lmarena}
	\begin{tabularx}{\textwidth}{LlLL}
		\toprule
		\textbf{\# 
} &\textbf{LLM Evaluators} & \textbf{Organization} & \textbf{License} \\
		\midrule
	   1 & claude-sonnet-4-5-20250929& Anthropic& Proprietary \\
        \midrule
        2 & Gemini-2.5-Flash-Preview-04-17& Google & Proprietary \\
        \midrule
        3 & o3-2025-04-16& OpenAI & Proprietary \\
		\midrule
        4 & DeepSeek-V3-0324& DeepSeek & MIT\\
		\midrule
        5 & GPT-4.1-2025-04-14& OpenAI  & Proprietary \\
		\bottomrule
	\end{tabularx}				
\end{table}
\vspace{-6pt}

\begin{table}[H]
	\small

\caption{GrAImes enthusiast evaluation: LLMs used as judges with the enthusiast prompt and responses compared against human literary~enthusiasts.}
\label{table:llms_enths_lmarena}
		\begin{tabularx}{\textwidth}{LlLl}
		\toprule

		\textbf{\#} 
            & \textbf{LLM Evaluators} & \textbf{Organization} & \textbf{License} \\
		\midrule
		1 & gemini-2.5-pro & Google & Proprietary \\
		\midrule
		2 & chatgpt-4o-latest-20250326 & OpenAI & Proprietary \\
		\midrule
		3 & claude-opus-4-20250514 & Anthropic & Proprietary \\
		\midrule
		4 & grok-4-0709 & xAI & Proprietary \\
		\midrule
		5 & gemini-2.5-flash & Google & Proprietary \\
		\midrule
		6 & gemini-2.5-flash-lite-preview-06-17-thinking & Google & Proprietary \\
		\midrule
		7 & gpt-4.1-2025-04-14 & OpenAI & Proprietary \\
		\midrule
		8 & o3-2025-04-16 & OpenAI & Proprietary \\
\midrule
9 & grok-3-preview-02-24 & xAI & Proprietary \\
\midrule
10 & deepseek-v3-0324 & DeepSeek & MIT \\

\midrule
11 & deepseek-r1-0528 & DeepSeek & MIT \\
\midrule
12 & claude-sonnet-4-20250514-thinking-32k & Anthropic & Proprietary \\
\midrule
13 & kimi-k2-0711-preview & Moonshot & Modified MIT \\
\midrule
14 & hunyuan-turbos-20250416 & Tencent & Proprietary \\
\midrule
15 & qwen3-235b-a22b-no-thinking & Alibaba & Apache 2.0 \\
\midrule
16 & mistral-medium-2505 & Mistral & Proprietary \\
\bottomrule
	\end{tabularx}				
\end{table}
\vspace{-6pt}

\subsubsection{Experiment
 Using TTCW Data with GrAImes to Compare Two Evaluation~Protocols}
 
    We conducted 
 an experiment using the same short story data written by humans and published in The New Yorker and~the stories generated by GPT-3.5-turbo, GPT4, and Claude V1.3 from the TTCW~\citep{chakrabarty2024art} evaluation with three LLMs, as~the original authors did (see Table~\ref{table:llms_ttcw_graimes_lmarena}). Because we could not use the same LLM versions, we selected the best-ranked ones from the same LLM models in the LMArena Creative Writing Leaderboard as of August/2025.

\begin{table}[H]
	\small
\caption{TTCW evaluation: list of LLMs used as judges with GrAIMes, the~expert prompt, and the short stories from the TTCW~experiment.}
    \label{table:llms_ttcw_graimes_lmarena}
	\begin{tabular}{p{1.5cm}  p{5.5cm} p{3cm} p{2.2cm}}
		\toprule

		\textbf{\#}
    & \textbf{LLM Evaluators} & \textbf{Organization} & \textbf{License} \\
		\midrule

1 &claude-opus-4-1-20250805&	Anthropic&	Proprietary\\ \midrule

2 &chatgpt-4o-latest-20250326&	OpenAI&	Proprietary\\		\midrule

3 &gpt-5-high&	OpenAI&	Proprietary\\		\bottomrule

	\end{tabular}				
\end{table}

\section[\appendixname~\thesection]        {Verbatim Prompts}
\label{Verbatim_prompts}
        {
        \subsection{Expert Prompt (Translated from Spanish) 
} \label{label:exp_prompt} Act as 
 a literary expert and evaluate the following piece of microfiction. For~your analysis, consider critical theories and approaches, intrinsic elements of the text, narratology, genre and subgenre, stylistics, intertextuality, symbolism, and~motifs. Your goal is to technically evaluate the text and determine its publication potential. Only answer the following questions: }
	
    {
	\subsection{Enthusiast Prompt (Translated from Spanish)}
    \label{label:enth_prompt} 

	In this task, the~objective is to evaluate the following microfiction texts and thereby determine their literary quality, and~in particular whether or not they deserve to be published. The~evaluation must be carried out from the perspective and literary background of a literature enthusiast and intensive reader that includes the work of authors such as Julio Cortázar, Augusto Monterroso, Ana María Shua, Juan Rulfo, Jorge Luis Borges and Roberto Bolaño. Just answer the following questions:}

{
Given the profiles used in the evaluation of TTCW~\citep{chakrabarty2024art} to which the answers of the LLMs were compared, we designed a prompt that covers the evaluators' backgrounds and~professions.

\subsection{Prompt}

As a literary expert with experience in teaching, editing, and~holding a Master of Fine Arts degree, I ask you to evaluate the following short story based on your knowledge of originality and creativity, theme and relevance, characterization, narrative structure and pacing, language and style, setting and atmosphere, and~marketability. Your analysis will determine its suitability for publication. Please respond only to the questions provided. This is an open-ended response with a maximum of 100 words and a Likert scale of 1 to 5:}

\section[\appendixname~\thesection]
        {GrAImes Literary Evaluation Protocol}\label{app3}
\vspace{-6pt}
        			\begin{table}[H]
				\footnotesize
				\caption{List of questions in the evaluation protocol provided to the evaluators tasked with assessing the literary, linguistic, and~editorial quality of the microfiction pieces. OA = Open Answer; Likert = Likert scale from 1 to~5.}
				\label{table:GraAImes_questions}
\begin{adjustwidth}{-\extralength}{0cm}

				\begin{tabularx}{\fulllength}{p{0.05cm} p{5cm} p{1cm} X}
					\midrule
					\multicolumn{4}{c}{GrAImes Evaluation Protocol Questions 
} \\
					\midrule
					\textbf{\#} & \textbf{Question} & \textbf{Answer} & \textbf{Description} \\
					\midrule
					\multicolumn{4}{c}{Story overview and Text Complexity} 
                     \\
					\midrule
					1 & What happens in the story? & OA & Evaluates how clearly the generated microfiction is understood by the~reader. \\
					\midrule
					2 & What is the theme? & OA & Assesses whether the text has a recognizable structure and can be associated with a specific~theme. \\
					\midrule
					3 & Does it propose other interpretations, in~addition to the literal one? & Likert & Evaluates the literary depth of the microfiction. A~text with multiple interpretations demonstrates greater literary~complexity. \\
					\midrule
					4 & If the above question was affirmative, Which interpretation is it? & OA & Explores whether the microfiction contains deeper literary elements such as metaphor, symbolism, or~allusion. \\
					\midrule
					\multicolumn{4}{c}{Technical Assessment}
                     \\
					\midrule
					5 & Is the story credible? & Likert & Measures how realistic and distinguishable the characters and events are within the~microfiction. \\
					\midrule
					6 & Does the text require your participation or cooperation to complete its form and meaning? & Likert & Assesses the complexity of the microfiction by determining the extent to which it involves the reader in constructing~meaning. \\
					\midrule
					7 & Does it propose a new perspective on reality? & Likert & Evaluates whether the microfiction immerses the reader in an alternate reality different from their~own. \\
					\midrule
					8 & Does it propose a new vision of the genre it uses? & Likert & Determines whether the microfiction offers a fresh approach to its literary~genre. \\
					\midrule
					9 & Does it give an original way of using the language? & Likert & Measures the creativity and uniqueness of the language used in the~microfiction. \\
					\midrule
					\multicolumn{4}{c}{Editorial/Commercial Quality}
                     \\
					\midrule
					10 & Does it remind you of another text or book you have read? & Likert & Assesses the relevance of the text and its similarities to existing works in the literary~market. \\
					\midrule
					11 & Would you like to read more texts like this? & Likert & Measures the appeal of the microfiction and its potential~marketability. \\
					\midrule
					12 & Would you recommend it? & Likert & Indicates whether the microfiction has an audience and whether readers might seek out more works by the~author. \\
					\midrule
					13 & Would you give it as a present? & Likert & Evaluates whether the microfiction holds enough literary or commercial value for readers to gift it to~others. \\
					\midrule
					14 & If the last answer was yes, to~whom would you give it as a present? & OA & Identifies the type of reader the evaluator believes would appreciate the~microfiction. \\
					\midrule
					15 & Can you think of a specific publisher that you think would publish a text like this? & OA & Assesses the commercial viability of the microfiction by determining if respondents associate it with a specific publishing~market. \\
					\midrule
				\end{tabularx}
\end{adjustwidth}
			\end{table}

			\begin{adjustwidth}{-\extralength}{0cm}
			\reftitle{References}

			\PublishersNote{}
			\end{adjustwidth}
	\end{document}